\documentclass{article}

\usepackage{arxiv}

\usepackage[utf8]{inputenc} 
\usepackage[T1]{fontenc}  
\usepackage{hyperref}      
\usepackage{url}           
\usepackage{booktabs}      
\usepackage{amsfonts}      
\usepackage{nicefrac}      
\usepackage{microtype}
\usepackage{lipsum}		
\usepackage{graphicx}
\usepackage{doi}
\usepackage{algorithm} 
\usepackage{algpseudocode}
\usepackage{multirow}
\usepackage{amsmath}
\usepackage{indentfirst}
\usepackage{braket}
\usepackage[title]{appendix}

\newcommand{\tmpframe}[1]{\fbox{#1}}

\title{Early-stage detection of cognitive impairment \\ by hybrid quantum-classical algorithm \\ using resting-state functional mri time-series \\}

\author{ 
        \hspace{1mm}{Junggu Choi}\\
	Yonsei Graduate program in Cognitive science\\
	Yonsei University\\
	Seoul, Republic of Korea \\
	\texttt{junggu.choi@yonsei.ac.kr} \\
	\And
	\hspace{1mm}{Tak Hur}\\
	Department of Statistics and Data Science\\
	Yonsei University\\
	Seoul, Republic of Korea \\
	\texttt{takh0404@yonsei.ac.kr} \\
	\AND
 	\hspace{1mm}{Daniel K. Park} \\
        Department of Applied Statistics\\	
        Department of Statistics and Data Science\\
	Yonsei University\\
	Seoul, Republic of Korea \\
	\texttt{dkd.park@yonsei.ac.kr} \\
        \And
	\hspace{1mm}{Na-Young Shin} \\
	Department of Radiology\\
	College of Medicine, Yonsei University\\
	Seoul, Republic of Korea \\
	\texttt{nyshin@yuhs.ac} \\
	\And
	\hspace{1mm}{Seung-Koo Lee} \\
	Department of Radiology\\
	College of Medicine, Yonsei University\\
	Seoul, Republic of Korea \\
	\texttt{slee@yuhs.ac} \\
	\And
	\hspace{1mm}{Hakbae Lee} \\
	Department of Applied Statistics\\
        Department of Statistics and Data Science\\
	Yonsei University\\
	Seoul, Republic of Korea \\
	\texttt{hblee@yonsei.ac.kr} \\
        \And
	\hspace{1mm}{Sanghoon Han}
        \thanks{Corresponding author} \\
	Department of Psychology\\
        Yonsei Graduate program in Cognitive science\\
	Yonsei University\\
	Seoul, Republic of Korea \\
	\texttt{sanghoon.han@yonsei.ac.kr}\\
}

\renewcommand{\shorttitle}{Choi \textit{et al.}, 2023}

\hypersetup{
pdftitle={Early-stage detection of cognitive impairment by hybrid quantum-classical algorithm using resting-state functional mri time-series},
pdfsubject={QML, EMCI},
pdfauthor={Choi. et al., 2023},
pdfkeywords={First keyword, Second keyword, More},
}

\begin{document}
\maketitle

\begin{abstract}
{Following the recent development of quantum machine learning techniques, the literature has reported several quantum machine learning algorithms for disease detection. This study explores the application of a hybrid quantum-classical algorithm for classifying region-of-interest time-series data obtained from resting-state functional magnetic resonance imaging in patients with early-stage cognitive impairment based on the importance of cognitive decline for dementia or aging. Classical one-dimensional convolutional layers are used together with quantum convolutional neural networks in our hybrid algorithm. In the classical simulation, the proposed hybrid algorithms showed higher balanced accuracies than classical convolutional neural networks under the similar training conditions. Moreover, a total of nine brain regions (left precentral gyrus, right superior temporal gyrus, left rolandic operculum, right rolandic operculum, left parahippocampus, right hippocampus, left medial frontal gyrus, right cerebellum crus, and cerebellar vermis) among 116 brain regions were found to be relatively effective brain regions for the classification based on the model performances. The associations of the selected nine regions with cognitive decline, as found in previous studies, were additionally validated through seed-based functional connectivity analysis. We confirmed both the improvement of model performance with the quantum convolutional neural network and neuroscientific validities of brain regions from our hybrid quantum-classical model.}
\end{abstract}

% keywords can be removed
\keywords{Resting-state functional magnetic resonance imaging \and Early mild cognitive impairment \and Quantum machine learning}

\section{Introduction}
%%% Contents of the introduction section.

%%% 1-1) 인지장애진단 연구의 중요성 소개
%% 인지장애 환자 초기 진단의 중요성
%% 기존에 활용되어 오던 진단법 및 데이터
%% 기존 진단법의 한계점(오분류율 관련)
{Diagnosis of early-stage cognitive impairment is critical for the tracking of patient-candidate groups that can progress to severe dementia. Previous studies have proposed methodologies for cognitive impairment diagnosis \cite{reference2, reference1, reference3}. Related studies and medical practices have mainly used questionnaire-based methods to detect cognitive decline \cite{reference4, reference5}. In Startin \textit{et al.}, the authors suggested modified questionnaires for the scaling of Down syndrome (DS) \cite{reference6}. Although several methods using questionnaires have been proposed in previous studies, some authors have pointed out their shortcomings. The response bias of self-report questionnaires in diagnosing Alzheimer's disease (AD) was shown by Hill \textit{et al.} \cite{reference8}. In their experimental results based on self-reported assessments, the authors found incorrect responses as well as inconsistencies therein. In addition, Brooks \textit{et al.} investigated the potential for the misclassification of mild cognitive impairment (MCI) based on the Wechsler memory scale test and a questionnaire \cite{reference9}. Furthermore, to minimize the misdiagnosis of MCI, Nikolai \textit{et al.} used computational tools to compare traditional criteria for MCI \cite{reference10}.}

%%% 1-2) 뉴로데이터(rs-fMRI) 기반 진단법 개발에 대한 최근 선행연구 소개
%% rs-fMRI 데이터 적용한 선행연구 소개
%% rs-fMRI 데이터를 활용할 때의 장점
{To address several limitations in methods utilizing the self-report questionnaire, other modalities, including neuroimaging, have been used in combination. Among various applicable modalities, a brain resting-state functional MRI (rs-fMRI) that collected to neural dynamics without any cognitive tasks have been applied for cognitive impairment detection. Wang \textit{et al.} validated the utility of the brain rs-fMRI for detecting vascular cognitive impairment \cite{reference11}. In addition, the authors compared fMRI responses from MCI-patient groups with those from healthy control groups in a picture-word pair paradigm to find differences in brain activation between the two groups \cite{reference12}. Zhang \textit{et al.} attempted to distinguish between early-MCI (EMCI) and late MCI groups using a functional brain network from the rs-fMRI \cite{reference13}. They found clear differences in the frequency bands from selected brain regions based on the MCI stage. Alterations of the default mode network and the quantification of these network changes were used as a biomarker for cognitive decline in Parkinson’s disease (PD) in their meta-analysis. Sun \textit{et al.} reviewed several previous studies on the analysis of single-channel time-series using electroencephalogram (EEG), magnetoencephalogram (MEG), and fMRI for AD patients \cite{reference14}. The authors suggested that a high spatial resolution of fMRI could provide more in-depth information for AD brain dysfunction than both EEG and MEG.}

%%% 1-3) 기존 데이터 분석법(주로 딥러닝)에 대한 소개
%%	기계학습을 이용한 선행연구 소개

{For multivariable analysis of the collected brain fMRI data, various data-analysis methods, including statistical modeling or machine learning (ML) algorithms, have been widely utilized. In particular, deep learning (DL) algorithms that show remarkable performance with many parameters have been widely used for classification or prediction tasks \cite{reference15, reference16}. Recent studies have also mainly utilized DL algorithms to classify brain fMRI data. Sarraf \textit{et al.} introduced convolutional neural network (CNN) based DL models to detect MCI and AD groups \cite{reference17}. The authors used preprocessed two-dimensional images from structural and functional MRI to train their model. Furthermore, Abdulaziz and Khan proposed a DL model-based pipeline for the multi-label classification of rs-fMRI data \cite{reference18}. In their study, a total of six labels of brain functional connectivities from each stage of AD patient (i.e., normal, significant memory concern, EMCI, MCI, late MCI, and AD) were classified. Moreover, high-order functional connectivity, which is calculated from the post-processing of the brain functional connectivity, is applied to classify the EMCI stage \cite{reference19}. The authors demonstrated the effectiveness of post-processed functional connectivity such as high-order and time-varying connectivity in EMCI diagnosis.}

%%% 1-4) 양자기계학습 소개
%%	양자기계학습을 활용한 선행연구 소개
%%	(기존 기계학습 방법론의 단점을 지적하며 소개할 것인가?)
%% -아래는 새로이 추가된 부분-
%% -> 언급가능한 기존 기계학습 방법론의 단점들 (양자머신러닝과 비교할 때의)
%%    1) model의 overparameterization? (GPT 계열 model과 같이)
%%    2) model training 시에 필요한 컴퓨팅 리소스가 너무 많이 필요함?

{Studies have reported several drawbacks of the DL algorithm. For example, Thompson \textit{et al.} discussed possible constraints related to the computational burden of DL models \cite{reference20}. They pointed out that the scale of model architecture such as the number of parameters could negatively impact a range of applications. In response to these challenges, recently, quantum machine learning (QML) has emerged as a novel approach for data analysis, motivated by the computational advantages of quantum computing. It aims to overcome the limitations inherent in its classical counterparts  \cite{reference21}. In particular, QML algorithms have made significant strides in classification tasks, showcasing their potential to outperform classical methods in terms of runtime efficiency, trainability, model capacity, and prediction accuracy \cite{reference22, reference23, reference24, reference25}. QML algorithms are also widely utilized for disease detection. Maheshwari \textit{et al.} applied optimized quantum support vector machine (OQSVM) and hybrid quantum multi-layer perceptron (HQMLP) to electronic healthcare records (EHR), which is one of the well-known structured datasets in the medical domain for ischemic heart-disease classification \cite{reference26}. They achieved higher classification accuracy compared with that achieved using classical SVM and MLP. Furthermore, the authors focused on the advantage of QML algorithms in terms of the efficiency of the computation time. In addition, Garg \textit{et al.} used quantum support vector machine (QSVM) models to recognize emotions from EEG signals \cite{reference27}. Several features including power spectral density (PSD) features were calculated from EEG signals. The authors found that QSVM models trained using EEG features showed improved classification performance compared with conventional SVM models. Furthermore, Felefly \textit{et al.} proposed a quantum neural network-based detection framework for solitary large-brain metastases and high-grade gliomas, which are difficult to differentiate on MRI \cite{reference28}. Among a total of 1,813 features extracted from two-dimensional MRI images, 10 were selected as the most associated features using the D-Wave quantum annealer. The authors confirmed that a 2-qubit quantum neural network had algorithm performance comparable to those of dense neural network and extreme gradient boosting as the classical counterparts. Although various types of data such as EHR, EEG, and structural MRI are being analyzed using QML algorithms, studies on rs-fMRI datasets pertaining to neurodegenerative patients have been rare. Furthermore, previous QML studies for time-series analysis have only used extracted features and not the raw time-series data\cite{reference29, reference30}.}

%%% 1-5) 본 연구에서 하고자 하는 것

{In this study, we classified rs-fMRI datasets collected from healthy control and EMCI patient groups using the quantum-classical hybrid ML algorithm. The rs-fMRI dataset for the two groups was preprocessed to ROI time-series to apply it to our algorithms as a one-dimensional time-series, as opposed to the widely used two-dimensional image. Time-series data on 116 ROIs were compared to examine their relative importance for the detection of EMCI based on the classification performance. The hybrid algorithm utilized a quantum convolutional neural network (QCNN) on a segment of the time-series data, while employing a classical convolutional neural network (CNN) without pooling layers on the remaining portion for local feature extraction and dimensionality reduction. Subsequently, the output features of these neural networks were combined and fed into a classical neural network to produce the final output. The QCNN was applied only to a portion of the data to ensure compatibility with Noisy Intermediate-Scale Quantum (NISQ) computers, which are becoming increasingly accessible \cite{preskill2018quantum}. In NISQ computers, the size of quantum circuits that can be reliably manipulated is limited due to noise and imperfections. This circumstance motivated the development of quantum-classical hybrid algorithms that rely on compact quantum models. The utilization of QCNN is motivated by the strengths outlined in the following: First, variational quantum algorithm (VQA) structures of the QCNN are relatively easy to implement in NISQ devices. Their shallow circuit depths and the utilizing of only nearest neighboring two qubit gates can offer advantages to the implementation in real quantum hardware with limited connectivity. Second, the QCNN have small generalization error upper bound by their small number of parameters. Caro \textit{et al} showed that the upper bound of the generalization error scales with number of parameters in QML algorithms \cite{reference32}. Since the QCNN has small number of parameters (the number of parameters grows as $O(log(n))$ for $n$ input qubits), it will have small generalization error upper bound. Third, it is known that the QCNN avoids the barren plateau problem, which is a vanishing gradient problem that hinders the optimization in QML. \cite{reference33}. Finally, the QCNN has shown good performances in classical and quantum tasks in previous studies \cite{reference34, reference35,hur2023neural}. Model performances were evaluated in simulations with several experimental conditions regarding the number of quantum replacements and 116 brain regions. The proposed model’s performance was validated through comparisons with classical baseline models. Our study has following strengths:}
\begin{itemize}
 \item We proposed a hybrid quantum-classical ML algorithm with QCNNs and classical CNN to classify ROI time-series signals from rs-fMRI datasets for healthy and EMCI groups.
	
 \item The hybrid algorithm outperformed its classical baseline-model in our simulation results. 
        
 \item Brain regions that showed larger improvements of the classification performance in the hybrid model have been validated by previous studies and additional seed-based functional connectivity (SBFC) analysis. 
\end{itemize}

{The remainder of this paper is organized as follows: Section 2 introduces the research scheme with descriptions of the rs-fMRI dataset, hybrid quantum-classical algorithm, baseline model, and evaluation methods for the classification performance of the algorithms. Section 3 describes the experimental results, including the classification performances of the algorithms and validation analysis regarding the brain-region selection based on the classification performance. Section 4 discusses the performance of the proposed hybrid algorithm for EMCI classification and brain regions related to the highest classification performance in previous studies. The conclusions, strengths, and limitations of this study are presented in Section 5.}

\section{Methods}
\begin{figure}[htb!]
\caption{The research scheme of this study.}
\centering
\tmpframe{\includegraphics[width=0.99\textwidth]{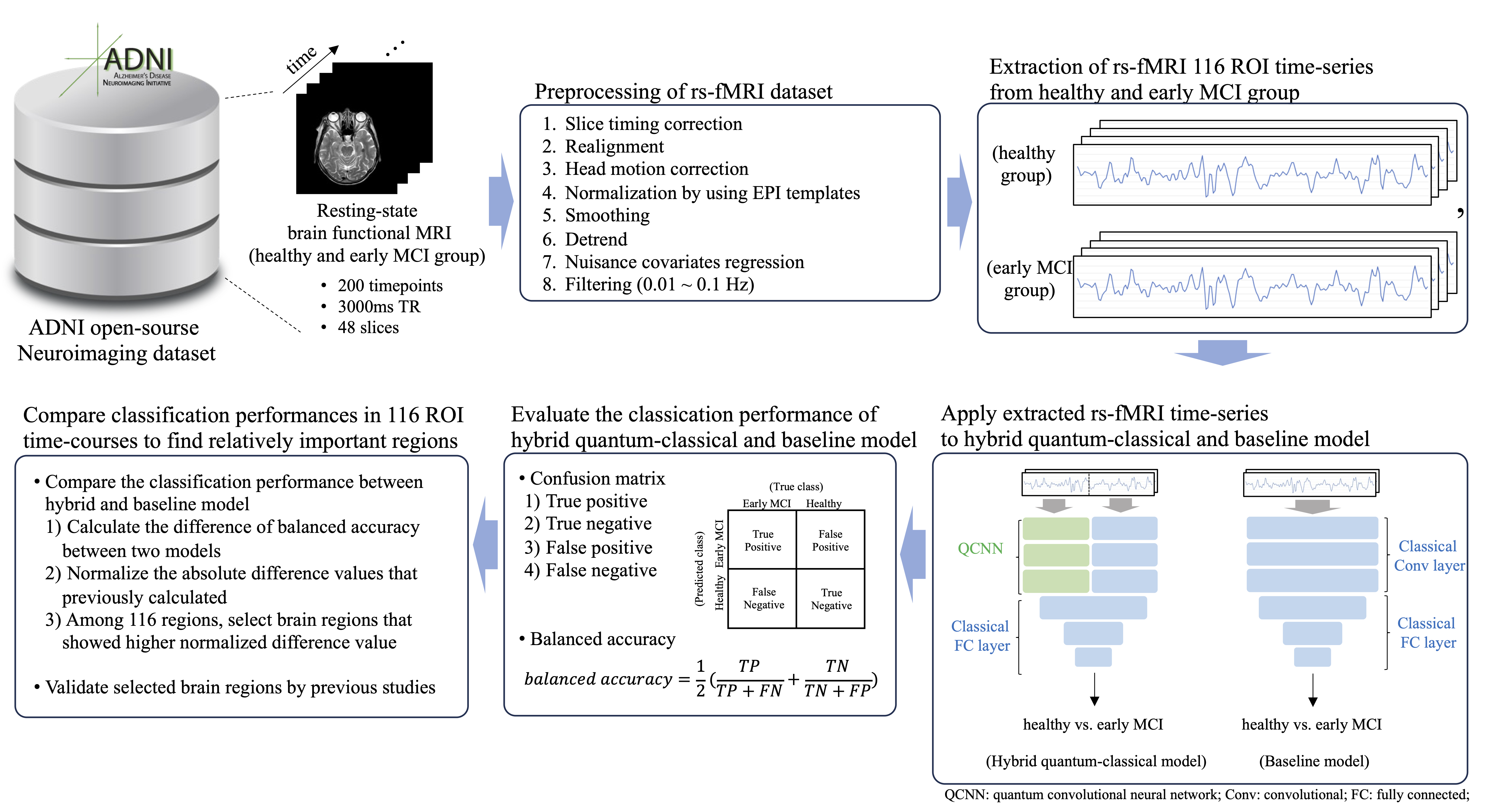}}
\label{fig:figure1}
\end{figure}

{An overview of this study is depicted in Figure \ref{fig:figure1}. The study comprises six main steps: 1) the collection of the rs-fMRI datasets measured from healthy and EMCI groups; 2) preprocessing the collected rs-fMRI datasets; 3) extracting the rs-fMRI region-of-interest (ROI) time-series from the preprocessed dataset; 4) training the hybrid quantum-classical and baseline models using the extracted time-series data; 5) evaluating the classification performances of the trained hybrid and baseline models; and 6) validating the importance of brain regions in model performances in each ROI condition.}

\subsection{Data source}
{The rs-fMRI data utilized in our research were sourced from the Alzheimer's Disease Neuroimaging Initiative (ADNI), which serves as a multi-site longitudinal-data repository \cite{reference36}. This initiative enables researchers to access publicly accessible data after obtaining approval, thereby significantly contributing to the advancement of research on AD. Within this repository, a vast collection of medical images, including MRI and Positron emission tomography (PET) scans, is available, alongside clinical, genomic, and biomarker data with the detailed five AD stages (i.e., healthy control, EMCI, MCI, late MCI, and AD). From several internal cohorts in ADNI (ADNI1, ADNI2, ADNIGO, and ADNI3), our study focused on the initial rs-fMRI images acquired during the medical follow-up of 365 individuals, encompassing 93 EMCI patients and 272 healthy individuals from all cohorts (male: 180, female: 185 / age: 74.75 $\pm$ 7.86 / weight: 75.77 $\pm$ 18.36).}

\subsection{MRI acquisition and preprocessing}
{In ADNI, all MRI data were obtained on 3T scanners based on the unified scanning protocols (detailed MR protocols and parameters are reported on: http://adni.loni.usc.edu/methods/mri-tool/mri-acquisition/). Because there are some differences in MR parameter between internal cohorts, rs-fMRI images were selected using the same three protocol conditions (200 timepoints, TR: 3000ms, and 48 slices). Consequently, the rs-fMRI images in ADNI2, ADNIGO, and ADNI3 were utilized in this study.}

{We conducted the preprocessing on all the rs-fMRI images stored in the Digital Imaging and Communications in Medicine (DICOM) format using the DPABI (toolbox for data processing \& analysis for brain imaging) Toolkit \cite{reference37}. This preprocessing workflow encompassed several important steps: conversion from DICOM to Neuroimaging Informatics Technology Initiative (NIFTI) format, slice-timing correction, realignment, head-motion correction, normalization to the MNI standard space, smoothing, detrending, and filtering (from 0.01 Hz to 0.1 Hz). Subsequently, we employed the automatic anatomical labeling (AAL) atlas, which divides the human brain into 116 distinct regions, to extract time-series data from these regions \cite{reference38}. Consequently, we obtained 116 ROI time-courses in the rs-fMRI datasets measured from healthy and EMCI groups.}

{Using the calculated 116 ROI time-series, we constructed final datasets to train and evaluate the hybrid quantum-classical and baseline models. The final datasets, $D = \{(X_{1},Y_{1}),(X_{2},Y_{2}),...(X_{N-1},Y_{N-1}),(X_{N},Y_{N})\}$, included a collection of pairs, $(X_i,Y_i)$, where $X_i$ indicates a univariate time-series with 140 data points ($X_i = [x_{1},x_{2},...,x_{139}, x_{140}]$) and $Y_i$ represents a class-label vector ($Y_i=0$ : healthy group and $Y_i=1$ : EMCI group). The dimension of the final dataset was (790, 141) (790 represents the number of rows and 141 indicates the number of columns). Among the 790 rows in the final dataset, the numbers of rows for the healthy and EMCI groups were 483 and 307, respectively.}

\subsection{Hybrid quantum-classical classification model}
\subsubsection{Model architecture}
%% 하이브리드 모델의 고전 알고리듬 파트와 양자 알고리듬 파트에서 사용된 알고리듬 소개와 선택함에 있어 이유가 되었던 선행연구 소개
{The hybrid quantum-classical model for the detection of EMCI was devised as follows. As the classical parts, we utilized one-dimensional CNN (1D CNN). This model was chosen based on the reported latent feature extraction performance of these algorithms from time-series data in previous studies. Among the various time-series data, the utility of 1D CNN for feature extraction from bio-signals such as electrocardiogram (ECG) or EEG was investigated. Jang \textit{et al.} applied 1D CNN layer architecture in an autoencoder model to extract latent features from actigraphy signal datasets in a denoising task \cite{reference39}. Furthermore, kang \textit{et al.} conducted a photoplethysmogram (PPG) and galvanic skin response (GSR) signal classification analysis with 1D CNN models \cite{reference40}. They found that the 1D CNN model showed improved classification performances using the limited computing resource and short input signals. From previous studies that showed the applicability of 1D CNN to the bio-signal classification, we determined that the classical parts in the hybrid model were set by the 1D CNN model structure. For the quantum part of the hybrid model, the QCNN proposed by Hur \textit{et al.} was utilized to investigate the effects of integrating quantum computing techniques with the 1D CNN model \cite{reference31}. The QCNN was selected because it has been shown to outperform its classical counterparts in the few-parameter regime \cite{reference31,reference34,reference35}. To test the change of the hybrid model performance with increasing number of QCNNs, we designed the small QCNN with four qubits in our hybrid model.}

%% 앞서 소개한 알고리듬을 기반으로 하이브리드 모델 내에서 데이터가 처리되는 플로우를 소개 
{Based on the aforementioned previous studies, our model architecture included the QCNN and classical 1D convolutional layers to evaluate influences of the quantum algorithm about model performance in the classification of fMRI time-series. In the two consecutive parts of our hybrid classification model (i.e., convolutional layer for the latent feature extraction and fully connected layer for the final classification), the classical 1D convolutional layers and QCNN, including quantum convolutional and quantum pooling layers, were used together in the convolutional layer. Therefore, the ROI signal was split into two parts for the QCNN and classical 1D convolutional layers. The latent features extracted from the classical 1D convolutional layers and measurement results (two probability values for 0 and 1) from the QCNN were concatenated as a single vector and applied to classify the two class labels in the classical fully connected layers.}

%% QCNN에서 적용된 amplitude encoding에 대한 설명
{In the QCNN of the hybrid model, the partial classical data in the ROI time-series were encoded to amplitudes of four qubits of the QCNN. The amplitude encoding applied in this study is among the general approaches to encoding classical data into a quantum state by normalizing classical input data as probability amplitudes of a quantum state. The formula for this encoding scheme is as follows: 

\begin{align}
    U_\phi(x) : x \in \mathbb{R}^N \rightarrow \ket{\phi(x)} = \frac{1}{\lVert x \rVert}\sum_{i=1}^N x_i\ket{i}
\end{align}

\noindent where $U_\phi(x) : x \in \mathbb{R}^N \rightarrow \ket{\phi(x)}$ indicate a unitary transformation (a quantum feature map) to make $U_\phi(x)\ket{0}^{\otimes n}=\ket{\phi(x)}$ from the initial state $\ket{0}^{\otimes n}$ of $n$ qubits. $x$ in the above formula represents the classical input sample ($x=(x_{1}, ..., x_{N})^T$) with dimension $N=2^n$ and $\ket{i}$ is the $i$th computational basis state in an $n$-qubit quantum state $\ket{\phi(x)}$. Therefore, sixteen classical data embedded to amplitudes of four qubits (i.e., $2^4=16$).} 

\begin{figure}[ht!]
\caption{The parameterized quantum circuit (PQC) used in the QCNN (the PQC a indicates the PQC for the quantum convolutional layer and the PQC b represents the PQC for the quantum pooling layer).}
\centering
\tmpframe{\includegraphics[width=10cm, height=5cm]{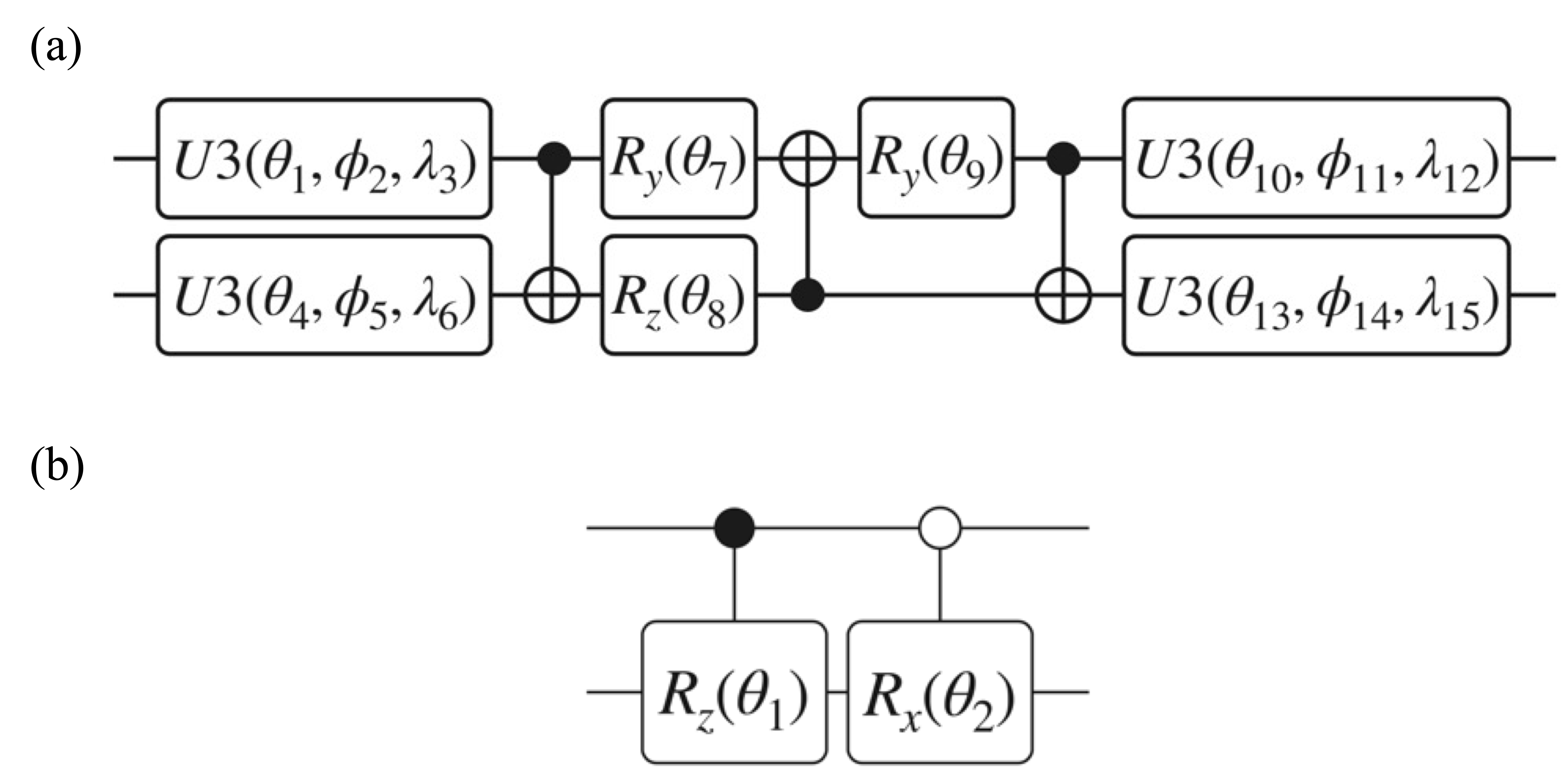}}
\label{fig:figure2}
\end{figure}

%% QCNN에서 사용된 quantum convolutional and quantum pooling layer에 대한 설명
{In the quantum convolutional layer, the two-qubit PQC that represented the parameterization of an \textit{SU}(4) gate with fifteen trainable parameters was applied \cite{reference43, reference44}. The PQC for the quantum convolutional layer is depicted in (a) of Figure \ref{fig:figure2}. In the quantum pooling layer, the approach is built upon two-qubit controlled rotation gates with two trainable parameters. The applied PQC in the quantum pooling layer are shown in (b) of Figure \ref{fig:figure2}. In two PQCs depicted in Figure \ref{fig:figure2}, $R_i(\theta)$ denotes a rotation gate around the $i$ axis of the Bloch sphere by an angle $\theta$ and $U3(\theta,\phi,\lambda)$ indicates an arbitrary single qubit gate ( $U3(\theta,\phi,\lambda) = R_{z}(\phi)R_{x}(-\pi/2)R_{z}(\theta)R_{x}(\pi/2)R_{z}(\lambda)$ ).}

\begin{figure}[hb!]
\caption{Hybrid model architecture example with a single QCNN}
\centering
\tmpframe{\includegraphics[width=0.99\textwidth]{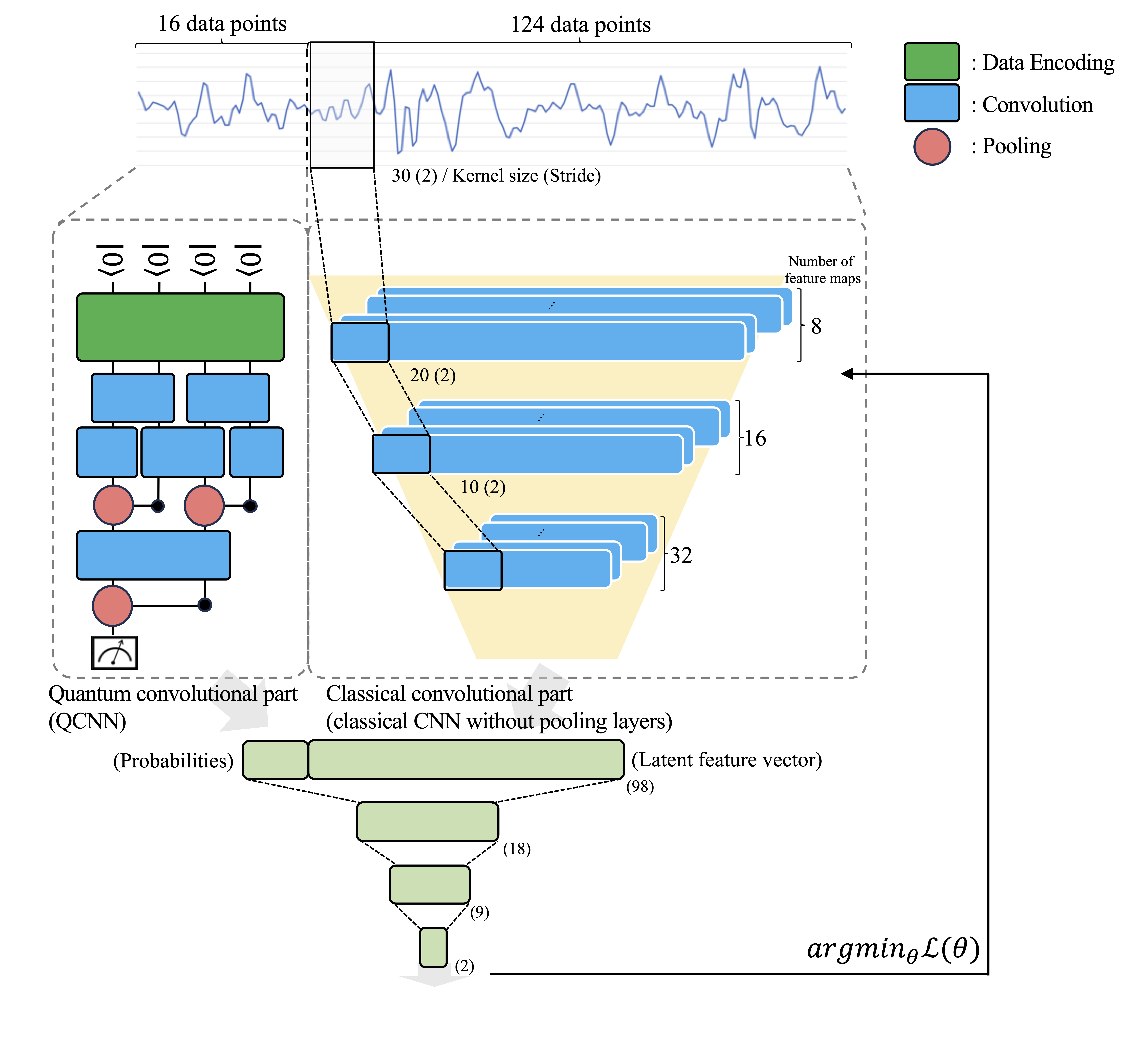}}
\label{fig:figure3}
\end{figure}

%% Classical CNN에서 처리되는 데이터에 대한 설명
{The remaining classical data in the ROI time-series were processed by classical 1D convolutional layers (total 140 data points - 16 data points for the single QCNN = 124 data points for the classical 1D convolutional layer). To compare model performances with varying numbers of the QCNN in the hybrid model, three different conditions for the number of QCNNs were tested (single, two, and four QCNNs). For the hybrid model with two QCNNs, 108 data points were applied as input data to the classic 1D convolutional layer (total 140 data points - 32 data points for two QCNNs = 108 classical data points to the classic 1D convolutional layer). Additionally, for the four QCNNs, 76 data points were processed by the classic 1D convolution layer (140 data points in total - 64 data points for the four QCNNs = 76 data points for the classic 1D convolution layer). An example of the hybrid-model architecture with a single QCNN is depicted in Figure \ref{fig:figure3}. Detailed descriptions of the remaining conditions, namely hybrid models with two and four QCNNs, can be found in Appendices A and B. In addition, the details of the proposed hybrid model with a single QCNN are listed in Table \ref{tab:table1}.}

\subsubsection{Simulation of hybrid quantum-classical model}
{Due to the difficulty of accessing real quantum hardware, we performed simulations on classical hardware to validate the classification performance of our hybrid quantum-classical algorithms under ideal conditions. We simulated the hybrid models and compared the performances with that of the classical 1D CNN model as a baseline model. The baseline model had a similar structure to that of the proposed hybrid quantum-classical algorithms. The baseline model’s architectures are listed in Table \ref{tab:table2}. We also calculated the number of parameters in the hybrid and baseline models to compare model performances corresponding the number of parameters. The number of trainable parameters in each model condition is shown in Table \ref{tab:table3}.}

{Moreover, the same training and evaluation conditions were used for the hybrid and baseline models. To optimize the trainable parameters of the hybrid and baseline models in the classification task, the cross-entropy loss function was applied. The optimizer for model training, batch size, the number of epochs, and learning rate was randomly selected (Adam optimizer, learning rate : 0.0001, and batch size : 1). Furthermore, class weight was used to reflect the imbalance of the rs-fMRI dataset in model training. All the models were validated in the 5-fold cross-validation with an 8:2 ratio of training to test datasets to evaluate the model performance based on all data. Additionally, because we need to consider true positive, true negative, false positive, and false negative together in the evaluation of disease classification model, balanced accuracy was applied in our study as an evaluation index to reflect the classification results including both the correctly classified and incorrectly classified samples. The formula for balanced accuracy is as follows:}

\begin{align}
    \textnormal{Balanced accuracy} = \frac{1}{2} \left( \frac{TP}{TP+FN} + \frac{TN}{TN+FP} \right)
\end{align}

\pagebreak
\begin{table}[ht]
\caption{The hybrid quantum-classical model architecture with a single QCNN}
\centering
\renewcommand{\arraystretch}{1.3}
\begin{tabular}{lllll}
\toprule[1.0pt]
\cmidrule(r){1-4}
\multicolumn{1}{c}{\multirow{2}{*}{Model structure}} & Input & Channel & Kernel & Output vectors \\
\cline{2-5}
\multicolumn{1}{c}{} &
  \begin{tabular}[c]{@{}l@{}}Dimension\\ (H $\times$ W $\times$ D)\end{tabular} &
  \begin{tabular}[c]{@{}l@{}}Input /\\ Output\end{tabular} &
  \begin{tabular}[c]{@{}l@{}}Kernel size /\\ Stride\end{tabular} &
  \begin{tabular}[c]{@{}l@{}}Dimension\\ (H $\times$ W $\times$ D)\end{tabular} \\
\midrule
Input & (1, 1, 140) &  &  & \\
\midrule
Partial input1 & (1, 1, 16) &  &  & \\
\midrule
Quantum Conv. layer1 & (1, 1, 16)  &   &  &   \\
Quantum Pooling layer1 & & & & \\
Quantum Conv. layer2 &  &  &  &  \\
Quantum Pooling layer2 & & & & (1, 1, 2) \\
\midrule
Partial input2 & (1, 1, 124) &  &  & \\
\midrule
Classical 1D Conv. layer1  & (1, 1, 124) & 1 / 8 & 30 / 2  & (1, 8, 48)  \\
Classical 1D Conv. layer2  & (1, 8, 48) & 8 / 16 & 20 / 2 & (1, 16, 15) \\
Classical 1D Conv. layer3  & (1, 16, 15) & 16 / 32 & 10 / 2  & (1, 32, 3) \\
Dropout 1D   &   &   &   & \\
\midrule
Concatenate outputs \\ from quantum and classical Conv layers & (1, 1, 98) & & & \\
\midrule
Classical fully connected layer1  & (1, 1, 98) & & & (1, 1, 18)\\
Classical fully connected layer2  & (1, 1, 18) & & & (1, 1, 9) \\
Classical fully connected layer3  & (1, 1, 9)  & & & (1, 1, 2)\\
\midrule
Output & & & & (1, 2) \\
\bottomrule[1.0pt]
\end{tabular}
\label{tab:table1}
\end{table} 

\begin{table}[ht]
\caption{Model architecture of the baseline model (classical CNN)}
\centering
\renewcommand{\arraystretch}{1.3}
\begin{tabular}{lllll}
\toprule[1.0pt]
\hline
\multirow{2}{*}{Model structure} & Input       & Channel & Kernel & Output vectors \\ \cline{2-5} 
 &
  \begin{tabular}[c]{@{}l@{}}Dimension\\ (H $\times$ W $\times$ D)\end{tabular} &
  \begin{tabular}[c]{@{}l@{}}Input/\\ Output\end{tabular} &
  \begin{tabular}[c]{@{}l@{}}Kernel size/\\ Stride\end{tabular} &
  \begin{tabular}[c]{@{}l@{}}Dimension\\ (H $\times$ W $\times$ D)\end{tabular} \\ \hline
Input                            & (1, 1, 140) &         &        &                \\ \hline
Classical 1D Conv. layer1        & (1, 1, 140) & 1 / 8   & 30 / 2 & (1, 8, 56)     \\
Classical 1D Conv. layer2        & (1, 8, 56)  & 8 / 16  & 20 / 2 & (1, 16, 19)    \\
Classical 1D Conv. layer3        & (1, 16, 19) & 16 / 32 & 10 / 2 & (1, 32, 5)     \\
Dropout 1D                       &             &         &        &                \\ \hline
Classical fully connected layer1 & (1, 1, 160)    &         &        & (1, 1, 18)        \\
Classical fully connected layer2 & (1, 1, 18)     &         &        & (1, 1, 9)         \\
Classical fully connected layer3 & (1, 1, 9)      &         &        & (1, 1, 2)         \\ \hline
Output                           &             &         &        & (1, 2)         \\ \hline
\bottomrule[1.0pt]
\end{tabular}
\label{tab:table2}
\end{table}

\pagebreak
\begin{table}[hp]
\caption{The number of trainable parameters in the hybrid and baseline models}
\centering
\renewcommand{\arraystretch}{1.3}
\begin{tabular}{ccc}
\toprule[1.0pt]
\hline
No. & Model conditions & \# of parameters \\ \hline
1   & Baseline model   & 11,605           \\ \hline
2 & \begin{tabular}[c]{@{}c@{}}Hybrid model cond. 1\\ (single quantum convolutional neural network)\end{tabular} & 9,983 \\ \hline
3 & \begin{tabular}[c]{@{}c@{}}Hybrid model cond. 2\\ (two quantum convolutional neural networks)\end{tabular}   & 8,901 \\ \hline
4 & \begin{tabular}[c]{@{}c@{}}Hybrid model cond. 3\\ (four quantum convolutional neural networks)\end{tabular}  & 4,177 \\ \hline
\bottomrule[1.0pt]
\end{tabular}
\label{tab:table3}
\end{table}

\pagebreak
\subsection{Additional analysis : seed-based functional connectivitiy (SBFC)}
{Following the comparisons of the classification performance between the hybrid and baseline models, we examined which brain region condition showed a larger difference in performance in the hybrid models than the baseline model. To examine the importance of the selected brain regions in the distinction between the two groups (healthy and EMCI groups), we additionally conducted SBFC analysis of selected brain regions.}

{Functional connectivity analysis in the seed-based approach requires seed regions with ROI time-series \cite{reference45, reference46}. The time-series is extracted from the seed region and the connectivity is then computed by temporally correlating seed time-series with the time-series of all other brain regions. From this analysis, we wanted to compare connectivity maps from the two groups with the same seeds based on the statistical tests. In addition, we assumed that if the differences in connectivity maps were in most brain regions with the same seed, the selected seed region was the main region to distinguish the two groups.}

\subsection{Tools}
{Hybrid quantum-classical and baseline algorithm codes were written by the Pennylane and Pytorch (Pennylane, version 0.25.1 ; Pytorch, version 1.7.1). Data preprocessing and visualization were built and operationalized using Python (version 3.7.1; scikit-learn, version 2.4.1).}

\section{Results}
\subsection{Classification performance in the hybrid and baseline models}
{To investigate the algorithm performance in the proposed hybrid quantum-classical models, we compared averaged balanced-accuracy values calculated from the hybrid and baseline models with the 5-fold cross-validation in the experimental conditions. In most experimental conditions with the 116 ROI time-series, the hybrid models showed higher classification performances than the baseline models (averaged balanced accuracy of the baseline model: 0.523; averaged balanced accuracy of the hybrid model with the single QCNN: 0.562; averaged balanced accuracy of the hybrid model with two QCNNs : 0.575; and averaged balanced accuracy of the hybrid model with four QCNNs : 0.581).}

\begin{figure}[hbt!]
\caption{The classification performance and the number of parameter changes of the hybrid model with the number of the QCNN (blue line with circles: the averaged balanced accuracy / red line with diamonds: the number of parameters).}
\centering
\tmpframe{\includegraphics[width=0.99\textwidth]{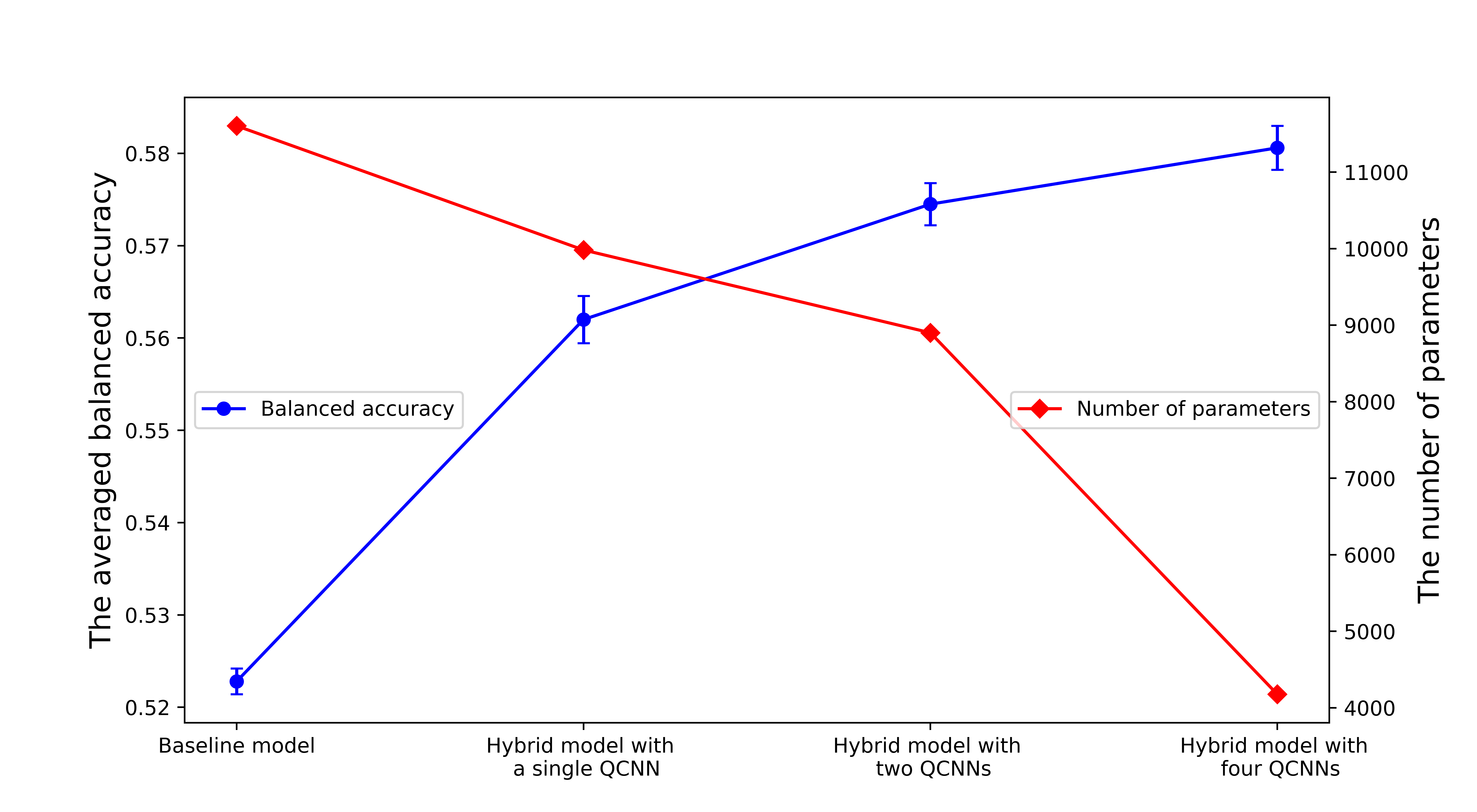}}
\label{fig:figure_added}
\end{figure}

{Moreover, in additional conditions pertaining to the number of QCNNs in the hybrid models, higher averaged balanced-accuracy values were found when using more QCNNs (averaged balanced accuracy of hybrid model with single QCNN : 0.562; averaged balanced accuracy of hybrid model with two QCNNs : 0.575; and averaged balanced accuracy of hybrid model with four QCNNs : 0.581). Furthermore, All differences between model conditions are statistically significant (\textit{p-value <0.05}). Detailed averaged balanced accuracies and normalized differences of balanced accuracy in the baseline and hybrid models with 116 experimental conditions are listed in Tables \ref{tab:table4} and \ref{tab:table5}. The performance changes of the hybrid model according to the number of parameters are summarized in Figure \ref{fig:figure_added}. Statistical test (independent two-sample t-test) results to validate statistical significance between model conditions are listed in Appendix C.}

\subsection{Comparisons of model performances in 116 brain regions}
{We evaluated the hybrid-model performances in a classical simulation based on comparisons with the baseline model using a quantitative indicator (averaged balanced accuracy). In addition, the hybrid-model performance must be investigated in terms of a qualitative evaluation (i.e., a function of the selected brain regions). To find ROI time-course conditions with the largest differences from the baseline model, three performance-difference categories of the baseline and hybrid models (hybrid with the single QCNN – baseline; hybrid with two QCNNs – baseline; and hybrid with four QCNNs - baseline) were averaged and normalized to single values. Based on the summarized single-difference values, 9 ROIs with the largest performance differences between the baseline and hybrid models (ROI 1, 84, 18, 17, 39, 38, 23, 92, and 110) were selected. We validated the selected 9 ROIs following interpretations for brain regions reported in previous studies. The selected 9 ROIs, with related information, are listed in Table \ref{tab:table6}.} 

\subsection{SBFC with selected ROIs}
\begin{figure}[hbt!]
\caption{The different brain regions with statistically significant differences in the SBFC maps between healthy and early-MCI groups (seed : ROI 1, 84, 18, 17, 39, and 38).}
\centering
\tmpframe{\includegraphics[width=0.99\textwidth]{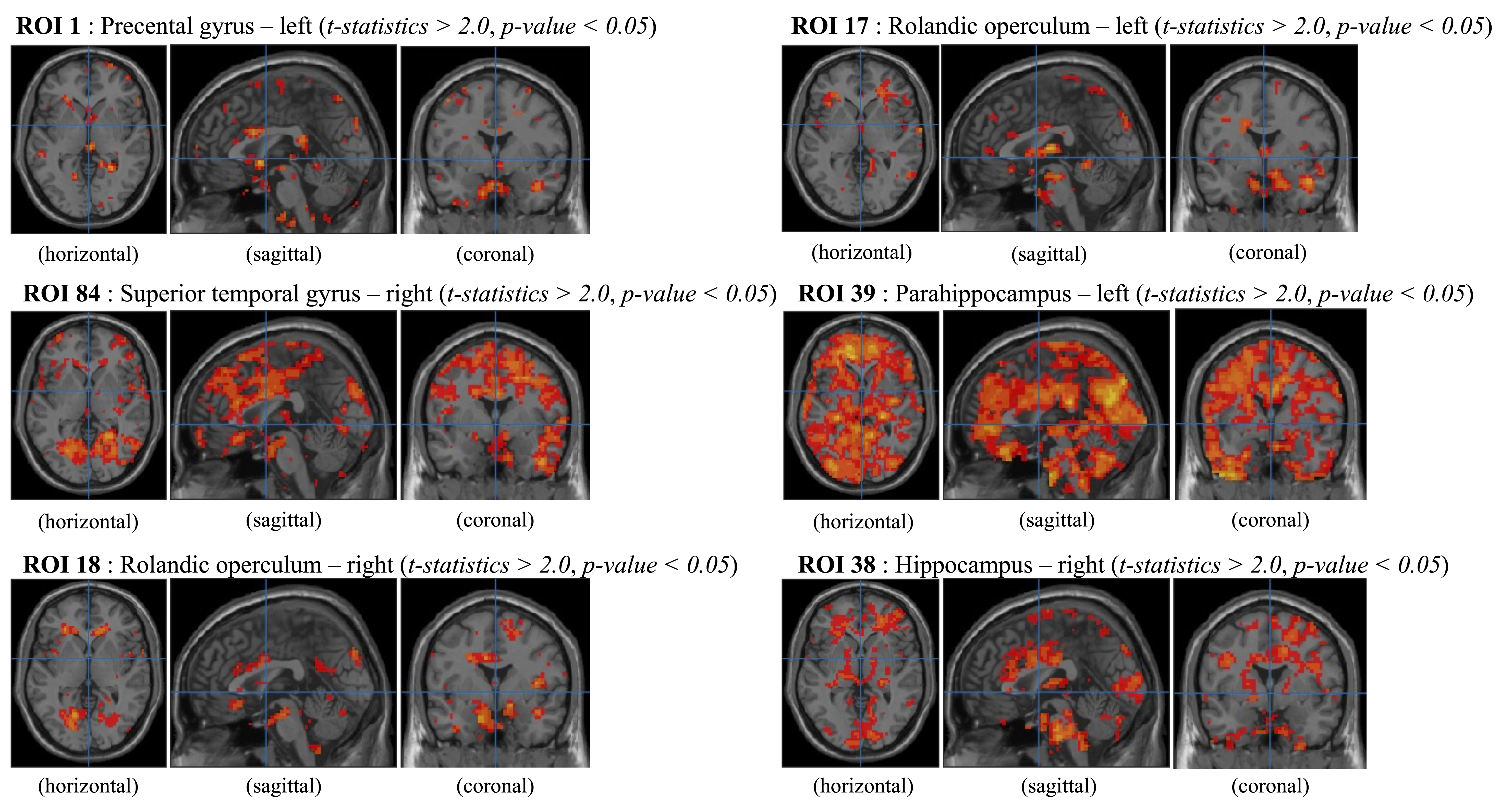}}
\label{fig:figure4}
\end{figure}

\begin{figure}[hbt!]
\caption{The different brain regions with statistically significant differences in the SBFC maps between healthy and early-MCI groups (seed : ROI 23, 92 , and 110).}
\centering
\tmpframe{\includegraphics[width=0.99\textwidth]{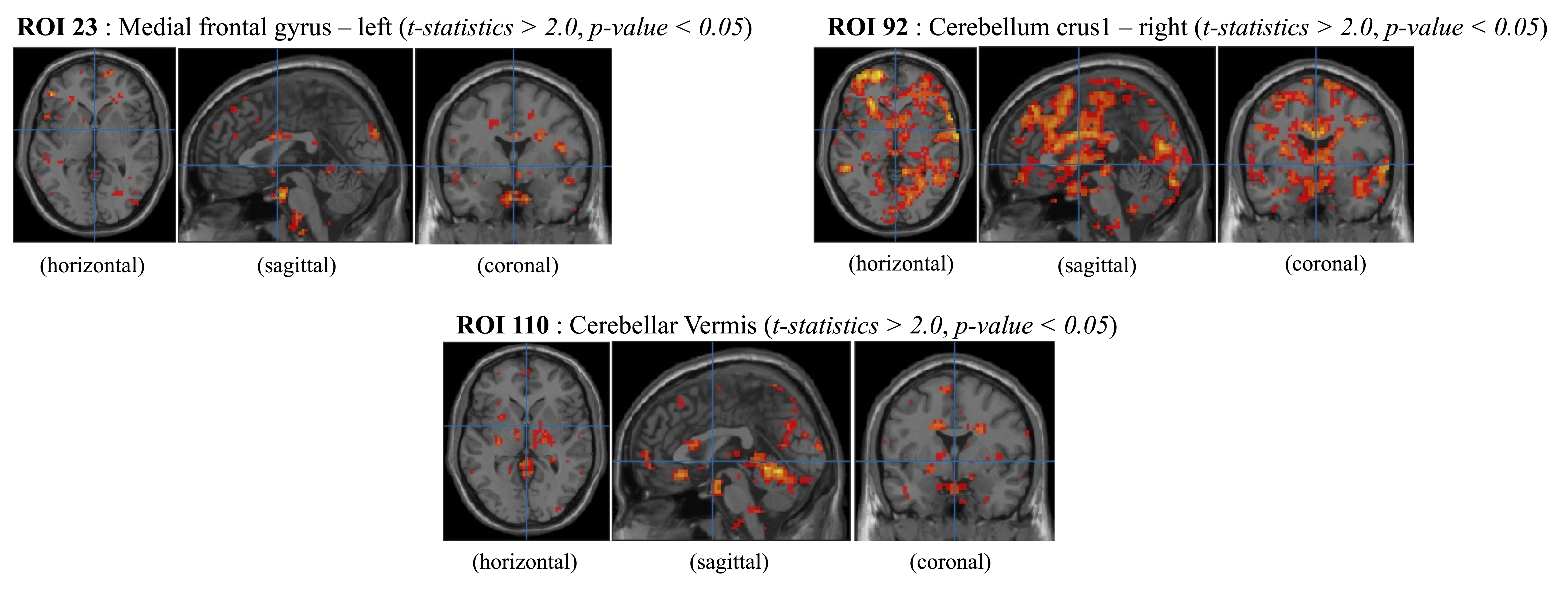}}
\label{fig:figure5}
\end{figure}

{Based on the aforementioned 9 ROIs, we additionally analyzed the importance of the selected brain regions in the distinction between healthy and EMCI groups. To investigate the centrality of the selected 9 brain regions in group comparisons, 9 brain regions were set as a seed in SBFC analysis. We compared brain regions with statistical significance in the difference between the two groups’ connectivity maps from the independent two-sample t-test. From the SBFC analysis results, statistically significant differences in connectivity maps were found in many regions. The different brain regions with statistically significant differences (\textit{t-statistics > 2.0} and \textit{p-value < 0.05}) in the SBFC maps are depicted in Figures \ref{fig:figure4} and \ref{fig:figure5}.}

\begin{figure}[hbtp!]
\caption{Summarized connections from the SBFC results with 9 seed regions}
\centering
\tmpframe{\includegraphics[width=\textwidth]{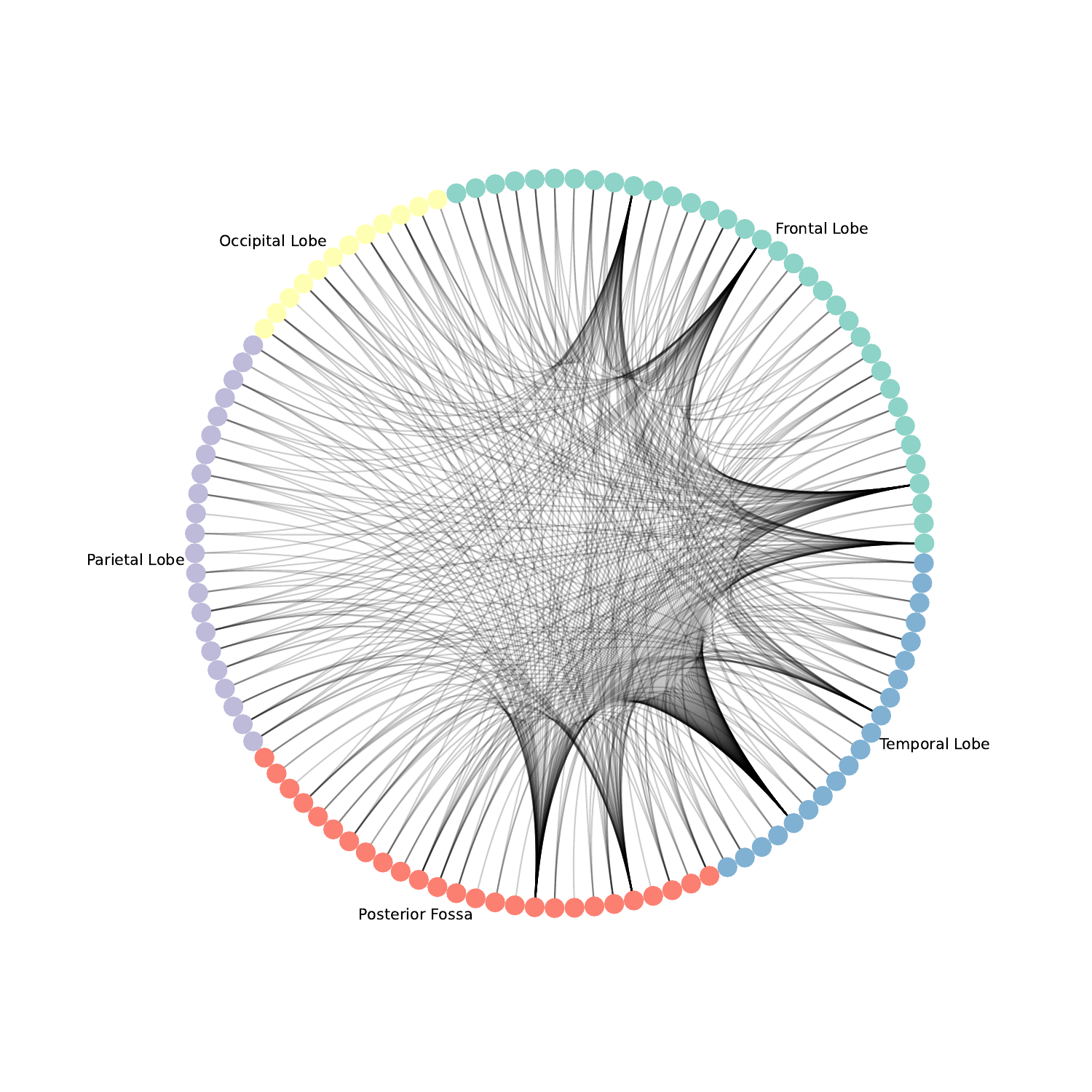}}
\label{fig:figure6}
\end{figure}

{Furthermore, brain regions can categorize into five higher categories (frontal lobe, temporal lobe, occipital lobe, parietal lobe, and posterior fossa) than the detailed 116 ROIs that were applied in this study as criteria. Therefore, to investigate which brain-region category among the 5 higher categories included the connections found, the connections between the brain regions in the SBFC results with 9 seed regions were summarized into a single circos plot. It was confirmed that brain regions belonging to the frontal lobe were selected the most as regions with the highest connectivity. The summarized connections based on the SBFC results are depicted in Figure \ref{fig:figure6}.}

\begin{table}[hbtp!]
\caption{Averaged balanced accuracy from the hybrid quantum-classical and baseline models trained by time-series from ROI 1 to 59.}
\centering
\small
\begin{tabular}{@{}cccccccc@{}}
\toprule
ROI &
  \begin{tabular}[c]{@{}c@{}}Baseline \\ model\end{tabular} &
  \begin{tabular}[c]{@{}c@{}}Hybrid \\ Cond.1\end{tabular} &
  \begin{tabular}[c]{@{}c@{}}Hybrid \\ Cond.2\end{tabular} &
  \begin{tabular}[c]{@{}c@{}}Hybrid \\ Cond.3\end{tabular} &
  \begin{tabular}[c]{@{}c@{}}Normalized \\ difference 1\\ (Hybrid1 - baseline)\end{tabular} &
  \begin{tabular}[c]{@{}c@{}}Normalized \\ difference 2\\ (Hybrid2 - baseline)\end{tabular} &
  \begin{tabular}[c]{@{}c@{}}Normalized \\ difference 3\\ (Hybrid3 - baseline)\end{tabular} \\ \midrule
1  & 0.510 & 0.591 & 0.632 & 0.615 & 0.918 & 0.977 & 1.000 \\
2  & 0.526 & 0.617 & 0.570 & 0.619 & 0.983 & 0.459 & 0.914 \\
3  & 0.530 & 0.489 & 0.552 & 0.573 & 0.108 & 0.320 & 0.584 \\
4  & 0.558 & 0.587 & 0.592 & 0.601 & 0.580 & 0.399 & 0.589 \\
5  & 0.512 & 0.600 & 0.508 & 0.529 & 0.967 & 0.148 & 0.411 \\
6  & 0.517 & 0.501 & 0.594 & 0.585 & 0.276 & 0.676 & 0.752 \\
7  & 0.537 & 0.536 & 0.511 & 0.515 & 0.373 & 0.000 & 0.151 \\
8  & 0.497 & 0.530 & 0.525 & 0.522 & 0.598 & 0.357 & 0.465 \\
9  & 0.524 & 0.530 & 0.578 & 0.576 & 0.420 & 0.526 & 0.643 \\
10 & 0.542 & 0.565 & 0.565 & 0.575 & 0.538 & 0.325 & 0.519 \\
11 & 0.522 & 0.521 & 0.572 & 0.584 & 0.374 & 0.500 & 0.710 \\
12 & 0.546 & 0.553 & 0.587 & 0.585 & 0.431 & 0.444 & 0.559 \\
13 & 0.530 & 0.544 & 0.590 & 0.570 & 0.480 & 0.567 & 0.565 \\
14 & 0.512 & 0.560 & 0.567 & 0.578 & 0.705 & 0.534 & 0.741 \\
15 & 0.513 & 0.561 & 0.587 & 0.570 & 0.705 & 0.659 & 0.676 \\
16 & 0.521 & 0.535 & 0.567 & 0.596 & 0.477 & 0.473 & 0.799 \\
17 & 0.486 & 0.549 & 0.597 & 0.577 & 0.797 & 0.898 & 0.902 \\
18 & 0.504 & 0.578 & 0.630 & 0.569 & 0.871 & 1.000 & 0.732 \\
19 & 0.528 & 0.582 & 0.502 & 0.501 & 0.741 & 0.003 & 0.123 \\
20 & 0.497 & 0.497 & 0.552 & 0.569 & 0.380 & 0.536 & 0.780 \\
21 & 0.522 & 0.545 & 0.550 & 0.595 & 0.535 & 0.359 & 0.782 \\
22 & 0.519 & 0.538 & 0.579 & 0.613 & 0.509 & 0.567 & 0.925 \\
23 & 0.502 & 0.568 & 0.585 & 0.596 & 0.823 & 0.722 & 0.923 \\
24 & 0.499 & 0.551 & 0.530 & 0.547 & 0.723 & 0.373 & 0.617 \\
25 & 0.522 & 0.491 & 0.569 & 0.527 & 0.175 & 0.484 & 0.331 \\
26 & 0.542 & 0.484 & 0.522 & 0.543 & 0.000 & 0.043 & 0.303 \\
27 & 0.504 & 0.472 & 0.536 & 0.545 & 0.173 & 0.382 & 0.570 \\
28 & 0.502 & 0.525 & 0.598 & 0.573 & 0.532 & 0.803 & 0.772 \\
29 & 0.540 & 0.560 & 0.602 & 0.594 & 0.517 & 0.578 & 0.658 \\
30 & 0.536 & 0.555 & 0.598 & 0.613 & 0.510 & 0.577 & 0.810 \\
31 & 0.528 & 0.566 & 0.571 & 0.547 & 0.634 & 0.455 & 0.423 \\
32 & 0.515 & 0.504 & 0.551 & 0.547 & 0.308 & 0.411 & 0.509 \\
33 & 0.513 & 0.538 & 0.597 & 0.591 & 0.548 & 0.722 & 0.813 \\
34 & 0.506 & 0.577 & 0.588 & 0.579 & 0.854 & 0.709 & 0.779 \\
35 & 0.510 & 0.561 & 0.579 & 0.551 & 0.718 & 0.625 & 0.571 \\
36 & 0.524 & 0.550 & 0.587 & 0.585 & 0.552 & 0.583 & 0.700 \\
37 & 0.512 & 0.559 & 0.596 & 0.579 & 0.695 & 0.726 & 0.741 \\
38 & 0.545 & 0.615 & 0.632 & 0.638 & 0.847 & 0.744 & 0.918 \\
39 & 0.511 & 0.602 & 0.600 & 0.595 & 0.982 & 0.753 & 0.852 \\
40 & 0.555 & 0.594 & 0.580 & 0.624 & 0.642 & 0.338 & 0.757 \\
41 & 0.540 & 0.574 & 0.565 & 0.559 & 0.605 & 0.339 & 0.421 \\
42 & 0.537 & 0.549 & 0.583 & 0.603 & 0.461 & 0.476 & 0.735 \\
43 & 0.517 & 0.558 & 0.564 & 0.583 & 0.655 & 0.482 & 0.740 \\
44 & 0.535 & 0.564 & 0.583 & 0.583 & 0.574 & 0.491 & 0.621 \\
45 & 0.512 & 0.568 & 0.602 & 0.587 & 0.753 & 0.768 & 0.796 \\
46 & 0.550 & 0.591 & 0.578 & 0.567 & 0.652 & 0.354 & 0.412 \\
47 & 0.514 & 0.554 & 0.578 & 0.570 & 0.645 & 0.588 & 0.667 \\
48 & 0.536 & 0.573 & 0.615 & 0.616 & 0.624 & 0.686 & 0.828 \\
49 & 0.542 & 0.594 & 0.574 & 0.615 & 0.723 & 0.383 & 0.783 \\
50 & 0.529 & 0.588 & 0.594 & 0.607 & 0.775 & 0.599 & 0.819 \\
51 & 0.509 & 0.587 & 0.568 & 0.598 & 0.900 & 0.559 & 0.890 \\
52 & 0.545 & 0.573 & 0.591 & 0.572 & 0.568 & 0.474 & 0.475 \\
53 & 0.539 & 0.571 & 0.568 & 0.591 & 0.594 & 0.360 & 0.642 \\
54 & 0.526 & 0.574 & 0.601 & 0.623 & 0.705 & 0.666 & 0.944 \\
55 & 0.511 & 0.578 & 0.579 & 0.580 & 0.827 & 0.619 & 0.759 \\
56 & 0.540 & 0.582 & 0.576 & 0.568 & 0.658 & 0.410 & 0.483 \\
57 & 0.536 & 0.585 & 0.528 & 0.491 & 0.711 & 0.122 & 0.000 \\
58 & 0.505 & 0.516 & 0.569 & 0.565 & 0.459 & 0.597 & 0.704 \\
59 & 0.553 & 0.557 & 0.565 & 0.546 & 0.412 & 0.254 & 0.252 \\ \bottomrule
\end{tabular}
\label{tab:table4}
\end{table}

\begin{table}[hbtp!]
\caption{Averaged balanced accuracy from the hybrid quantum-classical and baseline models trained by time-series from ROI 60 to 116.}
\centering
\small
\begin{tabular}{@{}cccccccc@{}}
\toprule
ROI &
  \begin{tabular}[c]{@{}c@{}}Baseline \\ model\end{tabular} &
  \begin{tabular}[c]{@{}c@{}}Hybrid \\ Cond.1\end{tabular} &
  \begin{tabular}[c]{@{}c@{}}Hybrid \\ Cond.2\end{tabular} &
  \begin{tabular}[c]{@{}c@{}}Hybrid \\ Cond.3\end{tabular} &
  \begin{tabular}[c]{@{}c@{}}Normalized \\ difference 1\\ (Hybrid1 - baseline)\end{tabular} &
  \begin{tabular}[c]{@{}c@{}}Normalized \\ difference 2\\ (Hybrid2 - baseline)\end{tabular} &
  \begin{tabular}[c]{@{}c@{}}Normalized \\ difference 3\\ (Hybrid3 - baseline)\end{tabular} \\ \midrule
60  & 0.501 & 0.558 & 0.599 & 0.572 & 0.761 & 0.817 & 0.767 \\
61  & 0.536 & 0.574 & 0.578 & 0.550 & 0.631 & 0.446 & 0.385 \\
62  & 0.526 & 0.567 & 0.563 & 0.576 & 0.658 & 0.417 & 0.635 \\
63  & 0.540 & 0.585 & 0.521 & 0.539 & 0.681 & 0.051 & 0.293 \\
64  & 0.519 & 0.538 & 0.583 & 0.563 & 0.510 & 0.595 & 0.596 \\
65  & 0.537 & 0.579 & 0.557 & 0.580 & 0.658 & 0.303 & 0.585 \\
66  & 0.505 & 0.581 & 0.566 & 0.573 & 0.887 & 0.573 & 0.750 \\
67  & 0.518 & 0.592 & 0.596 & 0.588 & 0.871 & 0.685 & 0.761 \\
68  & 0.537 & 0.584 & 0.569 & 0.581 & 0.691 & 0.382 & 0.586 \\
69  & 0.510 & 0.583 & 0.557 & 0.587 & 0.865 & 0.478 & 0.809 \\
70  & 0.524 & 0.544 & 0.569 & 0.572 & 0.515 & 0.470 & 0.617 \\
71  & 0.516 & 0.567 & 0.594 & 0.586 & 0.717 & 0.685 & 0.765 \\
72  & 0.537 & 0.593 & 0.603 & 0.571 & 0.758 & 0.605 & 0.525 \\
73  & 0.568 & 0.582 & 0.550 & 0.543 & 0.475 & 0.056 & 0.127 \\
74  & 0.515 & 0.578 & 0.538 & 0.580 & 0.796 & 0.319 & 0.730 \\
75  & 0.522 & 0.570 & 0.570 & 0.581 & 0.702 & 0.491 & 0.692 \\
76  & 0.523 & 0.591 & 0.582 & 0.608 & 0.838 & 0.561 & 0.867 \\
77  & 0.531 & 0.574 & 0.566 & 0.608 & 0.669 & 0.400 & 0.807 \\
78  & 0.527 & 0.571 & 0.553 & 0.575 & 0.672 & 0.340 & 0.619 \\
79  & 0.522 & 0.583 & 0.570 & 0.578 & 0.791 & 0.488 & 0.673 \\
80  & 0.528 & 0.579 & 0.596 & 0.585 & 0.722 & 0.618 & 0.678 \\
81  & 0.525 & 0.589 & 0.570 & 0.576 & 0.808 & 0.472 & 0.642 \\
82  & 0.504 & 0.567 & 0.557 & 0.594 & 0.799 & 0.520 & 0.897 \\
83  & 0.511 & 0.566 & 0.585 & 0.599 & 0.747 & 0.661 & 0.884 \\
84  & 0.499 & 0.591 & 0.596 & 0.590 & 1.000 & 0.809 & 0.910 \\
85  & 0.522 & 0.595 & 0.584 & 0.588 & 0.869 & 0.583 & 0.739 \\
86  & 0.518 & 0.569 & 0.561 & 0.594 & 0.719 & 0.450 & 0.800 \\
87  & 0.535 & 0.570 & 0.552 & 0.567 & 0.614 & 0.280 & 0.507 \\
88  & 0.523 & 0.580 & 0.609 & 0.603 & 0.762 & 0.735 & 0.829 \\
89  & 0.531 & 0.594 & 0.595 & 0.598 & 0.798 & 0.594 & 0.744 \\
90  & 0.532 & 0.610 & 0.546 & 0.613 & 0.897 & 0.260 & 0.834 \\
91  & 0.531 & 0.598 & 0.580 & 0.612 & 0.830 & 0.497 & 0.843 \\
92  & 0.521 & 0.590 & 0.592 & 0.620 & 0.841 & 0.641 & 0.959 \\
93  & 0.533 & 0.576 & 0.509 & 0.526 & 0.669 & 0.016 & 0.248 \\
94  & 0.523 & 0.563 & 0.563 & 0.597 & 0.650 & 0.438 & 0.793 \\
95  & 0.532 & 0.565 & 0.575 & 0.614 & 0.599 & 0.455 & 0.846 \\
96  & 0.509 & 0.571 & 0.568 & 0.583 & 0.789 & 0.557 & 0.790 \\
97  & 0.523 & 0.551 & 0.571 & 0.604 & 0.569 & 0.490 & 0.835 \\
98  & 0.520 & 0.571 & 0.582 & 0.587 & 0.726 & 0.582 & 0.750 \\
99  & 0.538 & 0.586 & 0.555 & 0.587 & 0.701 & 0.289 & 0.625 \\
100 & 0.528 & 0.571 & 0.582 & 0.598 & 0.669 & 0.529 & 0.763 \\
101 & 0.527 & 0.545 & 0.575 & 0.592 & 0.502 & 0.486 & 0.729 \\
102 & 0.498 & 0.528 & 0.587 & 0.578 & 0.582 & 0.762 & 0.837 \\
103 & 0.514 & 0.549 & 0.570 & 0.586 & 0.615 & 0.540 & 0.777 \\
104 & 0.509 & 0.538 & 0.573 & 0.587 & 0.580 & 0.599 & 0.823 \\
105 & 0.518 & 0.528 & 0.575 & 0.563 & 0.447 & 0.546 & 0.599 \\
106 & 0.501 & 0.539 & 0.567 & 0.586 & 0.632 & 0.602 & 0.862 \\
107 & 0.521 & 0.562 & 0.572 & 0.578 & 0.650 & 0.504 & 0.677 \\
108 & 0.529 & 0.561 & 0.571 & 0.595 & 0.593 & 0.448 & 0.735 \\
109 & 0.519 & 0.550 & 0.578 & 0.583 & 0.588 & 0.562 & 0.721 \\
110 & 0.493 & 0.536 & 0.589 & 0.592 & 0.667 & 0.808 & 0.957 \\
111 & 0.515 & 0.539 & 0.621 & 0.612 & 0.541 & 0.867 & 0.940 \\
112 & 0.535 & 0.560 & 0.598 & 0.591 & 0.547 & 0.587 & 0.674 \\
113 & 0.544 & 0.573 & 0.584 & 0.605 & 0.576 & 0.435 & 0.703 \\
114 & 0.518 & 0.548 & 0.590 & 0.557 & 0.581 & 0.646 & 0.561 \\
115 & 0.507 & 0.540 & 0.577 & 0.591 & 0.599 & 0.633 & 0.856 \\
116 & 0.512 & 0.551 & 0.593 & 0.597 & 0.639 & 0.709 & 0.868 \\ \bottomrule
\end{tabular}
\label{tab:table5}
\end{table}

\begin{table}[ht!]
\caption{Selected top 9 ROIs by averaged performance differences between the baseline and hybrid models.}
\centering
\small
\begin{tabular}{@{}ccccc@{}}
\toprule
Rank &
  \begin{tabular}[c]{@{}c@{}}Brain region\\ (abbreviation)\end{tabular} &
  \begin{tabular}[c]{@{}c@{}}Brain region\\ (full name)\end{tabular} &
  \begin{tabular}[c]{@{}c@{}}Average of \\ Norm. diff\end{tabular} &
  \begin{tabular}[c]{@{}c@{}}ROI \\ No.\end{tabular} \\ \midrule
1 & Precentral\_L        & Precentral gyrus                       & 0.965 & 1   \\
2 & Templ\_Pole\_Sup\_R  & Temporal pole: superior temporal gyrus & 0.906 & 84  \\
3 & Rolandic\_Oper\_R    & Rolandic operculum                     & 0.868 & 18  \\
4 & Rolandic\_Oper\_L    & Rolandic operculum                     & 0.866 & 17  \\
5 & ParaHippo\_L         & Parahippocampus                        & 0.862 & 39  \\
6 & Hippocampus\_R       & Hippocampus                            & 0.836 & 38  \\
7 & Frontl\_Sup\_Medl\_L & Medial frontal gyrus                   & 0.823 & 23  \\
8 & Cerebelm\_Crus1\_R   & Cerebelm crus                          & 0.814 & 92  \\
9 & Vermis\_3            & Cerebellar Vermis                      & 0.810 & 110 \\ \bottomrule
\end{tabular}
\label{tab:table6}
\end{table}

\clearpage
\section{Discussion}
%% 문단1: 연구결과에 대한 본격적인 해석 전에 간단히 연구과정에 대한 요약 
{In summary, we proposed a hybrid quantum-classical algorithm to classify patients' early-stage cognitive impairment based on rs-fMRI time-series datasets. Rs-fMRI datasets collected from healthy and EMCI groups in the ADNI dataset were analyzed. After preprocessing the datasets, raw rs-fMRI time-series were applied as input data to the proposed hybrid model. A total of 116 time-series on 116 brain regions were used to evaluate the hybrid model’s performance. In the hybrid model, both classical 1D convolutional layers and the QCNN was applied. To investigate the influences of the QCNN in terms of the model’s performance, the baseline model’s performances were compared under the same experimental conditions. Based on the model performance for each brain region, we examined the relative importance of brain region in this classification task. Furthermore, confirmed brain regions that were associated with higher classification performance were additionally validated through SBFC analysis. From this study, the following three main perspectives to interpret our experimental results are found: first, input data type from the rs-fMRI datasets; second, the proposed hybrid model’s architecture. Third, the brain region comparison from the model performance. We discuss these points as follows:}

%% 문단2: rs-fMRI 데이터분석에 있어 raw time-series 데이터를 그대로 사용한 점
{First, in the case of the input dataset for algorithm evaluation, most previous studies that analyzed fMRI datasets using ML models were based on post-processed datasets such as connectivity matrices or feature vectors from the connectivity matrix \cite{reference46, reference47}. In addition, to model fMRI time-series signals measured from task-based research (task fMRI), Huang \textit{et al.} applied raw time-series datasets to classical convolutional autoencoder models \cite{reference48}. They examined the application of raw time-course signals from fMRI to detect AD. Among possible approaches regarding the type of fMRI datasets, we selected the raw ROI time-course dataset as the model input. This is because examining the usabilities of raw fMRI time-series signals with improved classification performance on the hybrid approach with QCNN can facilitate dataset post-processing for further studies using fMRI datasets.}

%% 문단3: raw time-series 데이터를 기반으로 하였었던 본 연구의 hybrid 모델성능
{Second, associated with the use of raw time-series, we confirmed higher averaged balanced-accuracy values from our hybrid models than those obtained in the baseline models. Moreover, the hybrid models’ classification performance showed gradual improvements when the number of QCNN was increased (single-, two-, and four-QCNN conditions). Furthermore, we found that the number of trainable parameters in the hybrid models decreased relative to that in the baseline models. The baseline model included 11,065 parameters. Meanwhile, the hybrid models with the QCNN included 9,983 (single QCNN), 8,901 (two QCNNs), and 4,177 (four QCNNs) parameters. Based on our experimental results, we found that the hybrid algorithm’s classification performances can be improved based on the application of the QCNN along with the classical 1D convolutional layers.}

%% 문단4: 본 연구에서 제시한 hybrid 모델 구조의 특이한 점 (splitted input data / hybrid 모델 내 QCNN의 배치)
{Third, for the hybrid model structure with the QCNN, many studies have reported hybrid algorithms with sequential compositions of the PQC and classical DL algorithms. For example, Lockwood and Si investigated the performance of hybrid quantum-classical reinforcement learning algorithms in Atari game as a testbed \cite{reference49}. Their hybrid algorithms included QCNNs and classical double deep Q learning (DDQN) algorithms repeated in a single data flow. In addition, Houssein \textit{et al.} applied quantum convolutional layers consisted of PQCs in hybrid models to predict COVID-19 from chest X-ray images \cite{reference50}. These quantum convolutional layers was utilized to extract information from two-dimensional X-ray images. The output of quantum convolutional layers was sequentially inputted to classical neural networks. Unlike previous studies, we attempted a non-sequential arrangement of the QCNN and the classical CNN in our hybrid models. In this arrangement, partial informations in splited raw rs-fMRI signals were processed by the QCNN and classical 1D convolutional layers in the classical CNN model. The outputs from each QCNN and classical CNN were concatenated to single vectors as an input vector for classical fully connected layers. From our experiments, we checked that applications of the QCNN and classical CNN with a non-sequential composition in the hybrid model can improve the model performance in the time-series classification task.} 

%% 문단5: LSTM 기반의 모델과의 성능비교결과 소개
{Fourth, a recurrent neural network (RNN) structure has been widely utilized for temporal-data analysis. We conducted the additional experiment to verify the classification performance of the hybrid model with other baseline models with a classical long short-term memory (LSTM) algorithm. The baseline model including the classical LSTM algorithm showed the lower averaged balanced accuracy values than the baseline model which we originally applied in this work. From this result, we additionally confirmed the model performance improvement in our hybrid model from two different baseline models. Detailed experimental results are listed in Appendix D.}

%% 문단6: 뇌과학적 해석 (1) - 116개 뇌 영역 별로 모델성능을 기준으로 비교 시 확인된 결과 해석
{Finally, regarding the differences in model performance between the 116 brain ROI conditions, nine ROI conditions that showed the highest averaged balanced accuracy were selected. We considered that the selected nine ROIs had higher relative importance for the classification between EMCI and AD than other ROIs. Moreover, the importance of each ROI was compared with that in previous studies. For the selected nine ROIs, the association of the precentral gyrus (Precentral\char`_L / ROI 1) and hippocampus (Hippocampus\char`_R / ROI 38 and ParaHippo\char`_L / ROI 39) in functional connectivity on cognitive decline was confirmed by Han \textit{et al} \cite{reference53}. In addition, based on our experimental results, we reconfirmed the correlation between the superior temporal gyrus (Templ\char`_Pole\char`_Sup\char`_R / ROI 84) and MCI in patients found in previous studies in fMRI research\cite{reference54}. Furthermore, the relationships that were previously found between cognitive decline and the rolandic operculum (Rolandic\char`_Oper\char`_L, R / ROI 17, 18) and medial frontal gyrus (Frontl\char`_Sup\char`_Medl\char`_L / ROI 23) were identical to those found in our study \cite{reference55, reference56}. Furthermore, brain regions in the cerebellum have received attention as related regions for cognitive impairment in recent studies \cite{reference57,reference58}. We found that cerebellum crus 1 (Cerebelm\char`_Crus1\char`_R / ROI 92) and cerebellar vermis (Vermis\char`_3 / ROI 110) were important in detecting EMCI, as found in previous studies.}

%% 문단7: 뇌과학적 해석 (2) - 앞서 확인된 주요 뇌영역을 SBFC로 추가분석 시 확인된 결과 해석
{To validate the selected nine ROIs, we additionally conducted an SBFC analysis based on the selected ROIs as seed regions. Brain regions that showed a statistically significant difference between the two groups’ connectivity maps were organized to examine the influences of seed regions on the classification. Most of the brain regions in the 116 ROIs were found in the SBFC results. The frontal lobe region, which has already been proved to play a major role in cognitive impairment, was also identified in the summarized result \cite{reference59, reference60}. We concluded that these selected regions from our hybrid models played central roles in the classification.}

\section{Conclusion}
{In this study, we proposed hybrid quantum-classical ML algorithms for the classification of ROI time-series datasets from healthy and EMCI groups. Our research scheme found improved classification performances in hybrid models with fewer parameters than baseline models. From our experimental results, we found that the hybrid model with a higher number of the QCNN achieved higher classification performance than the classical CNN model (baseline model). Moreover, a total of nine ROI conditions that showed the largest differences between the hybrid and baseline models’ performances were selected from 116 ROI conditions. We validated that the selected nine ROIs were associated with cognitive decline among patient groups with cognitive impairment, as reported in previous studies. The selected nine ROIs were used as seed regions for an SBFC analysis to additionally confirm whether they were the main regions for the classification.}

{Our study has several strengths in various aspects. First, we proposed hybrid quantum-classical algorithms with the QCNN and classical CNN together. Second, our model showed higher classification performance with raw fMRI time-series signals from rs-fMRI datasets as input data without any additional post-process such as a connectivity matrix. Third, we validated that some ROI signals that showed a larger improvement in algorithm performance were related to those reported in previous studies. However, there are some limitations. First, only rs-fMRI time-series signal datasets were applied in this study. Therefore, various other dataset modalities (e.g., structural MRI or demographic variables) should be adopted to improve model performance. Second, the proposed hybrid quantum-classical algorithms were only examined by numerical simulations. To evaluate the robustness of our hybrid algorithm on the NISQ device, we need to examine the performance of the proposed algorithms using real quantum hardware. We plan to include other modalities in analysis and test our algorithm on real quantum hardware in future studies with same research theme.}

\section*{Acknowledgments}
This research was supported by Institute for Information \& communications Technology Promotion (IITP) grant funded by the Korea government (No. 2019-0-00003, Research and Development of Core technologies for Programming, Running, Implementing and Validating of Fault-Tolerant Quantum Computing System), the Yonsei University Research Fund of 2023 (2023-22-0072), and the National Research Foundation of Korea (Grant No. 2022M3E4A1074591).

\clearpage
\bibliographystyle{unsrt}
\bibliography{references}

\clearpage
\begin{appendices}
\counterwithin{figure}{section}

\section{Example of the hybrid quantum-classical models with two and four QCNNs.}

\begin{figure}[htp!]
\caption{The hybrid-model architecture example with two QCNNs.}
\centering
\tmpframe{\includegraphics[width=0.99\textwidth]{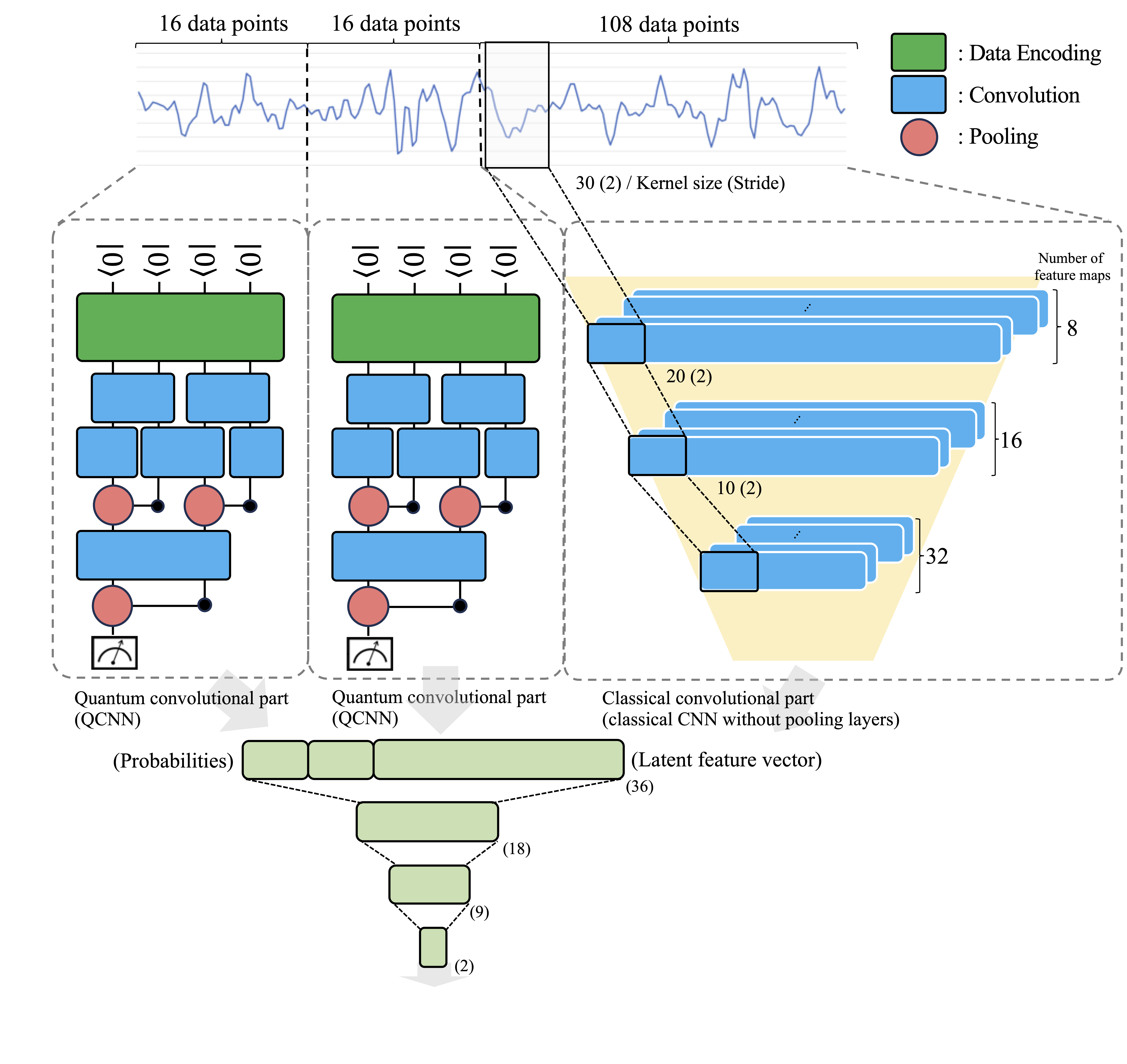}}
\label{fig:appendix_figure_a1}
\end{figure}

\begin{figure}[htp!]
\caption{The hybrid-model architecture example with four QCNNs.}
\centering
\tmpframe{\includegraphics[width=0.99\textwidth]{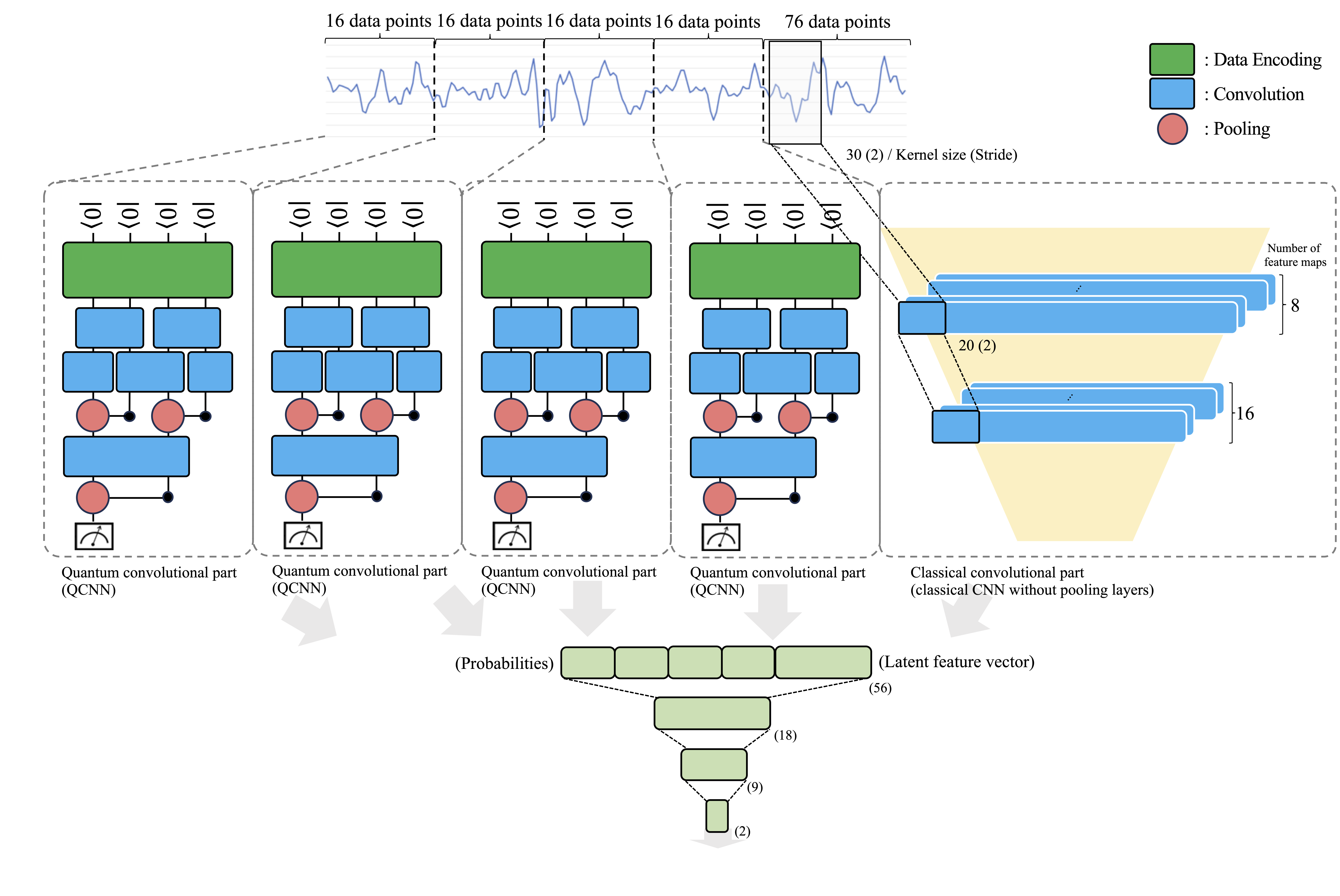}}
\label{fig:appendix_figure_a2}
\end{figure}

\clearpage
\section{Model architectures in the hybrid quantum-classical models with two and four QCNNs.}

\setcounter{table}{0}
\renewcommand{\thetable}{B\arabic{table}}

\begin{table}[htp!]
\caption{Model architecture in the hybrid quantum-classical model with two QCNNs}
\centering
\renewcommand{\arraystretch}{1.3}
\begin{tabular}{lllll}
\toprule[1.0pt]
\hline
\multirow{2}{*}{Model structure} & Input       & Channel & Kernel & Output vectors \\ \cline{2-5} 
 &
  \begin{tabular}[c]{@{}l@{}}Dimension\\ (H $\times$ W $\times$ D)\end{tabular} &
  \begin{tabular}[c]{@{}l@{}}Input /\\ Output\end{tabular} &
  \begin{tabular}[c]{@{}l@{}}Kenel size / \\ Stride\end{tabular} &
  \begin{tabular}[c]{@{}l@{}}Dimension\\ (H $\times$ W $\times$ D)\end{tabular} \\ \hline
Input                            & (1, 1, 140) &         &        &               \\ \hline
Partial input1                   & (1, 1, 16)  &         &        &               \\ \hline
Quantum Conv. layer1              & (1, 1, 16)  &         &        &               \\
Quantum Pooling layer1 & & & & \\
Quantum Conv. layer2              &             &         &        & \\ 
Quantum Pooling layer2 & & & & (1, 1, 2) \\ \hline
Partial input2                   & (1, 1, 16)  &         &        &               \\ \hline
Quantum Conv. layer1              & (1, 1, 16)  &         &        &               \\
Quantum Pooling layer1 & & & & \\
Quantum Conv. layer2              &             &         &        &      \\ 
Quantum Pooling layer2 & & & & (1, 1, 2) \\ \hline
Partial input3                   & (1, 1, 108) &         &        &               \\ \hline
Classical 1D Conv. layer1        & (1, 1, 108) & 1 / 8   & 30 / 2 & (1, 8, 40)    \\
Classical 1D Conv. layer2        & (1, 8, 40)  & 8 / 16  & 20 / 2 & (1, 16, 11)   \\
Classical 1D Conv. layer3        & (1, 16, 11) & 16 / 32 & 10 / 2 & (1, 32, 1)    \\ \hline
\begin{tabular}[c]{@{}l@{}}Concatenate outputs\\ from quantum and classical Conv layers\end{tabular} &
  (1, 1, 36) &
   &
   &
   \\ \hline
Classical fully connected layer1 & (1, 1, 36)  &         &        & (1, 1, 18)    \\
Classical fully connected layer2 & (1, 1, 18)  &         &        & (1, 1, 9)     \\
Classical fully connected layer3 & (1, 1, 9)   &         &        & (1, 1, 2)     \\ \hline
Output                           &             &         &        & (1, 2)        \\ \hline
\bottomrule[1.0pt]
\end{tabular}
\end{table}

\begin{table}[htp!]
\caption{Model architecture in the hybrid quantum-classical model with four QCNNs}
\centering
\renewcommand{\arraystretch}{1.3}
\begin{tabular}{lllll}
\toprule[1.0pt]
\hline
\multirow{2}{*}{Model structure} & Input       & Channel & Kernel & Output vectors \\ \cline{2-5} 
 &
  \begin{tabular}[c]{@{}l@{}}Dimension\\ (HxWxD)\end{tabular} &
  \begin{tabular}[c]{@{}l@{}}Input /\\ Output\end{tabular} &
  \begin{tabular}[c]{@{}l@{}}Kenel size / \\ Stride\end{tabular} &
  \begin{tabular}[c]{@{}l@{}}Dimension\\ (HxWxD)\end{tabular} \\ \hline
Input                            & (1, 1, 140) &         &        &               \\ \hline
Partial input1                   & (1, 1, 16)  &         &        &               \\ \hline
Quantum Conv. layer1             & (1, 1, 16)  &         &        &               \\
Quantum Pooling layer1 & & & & \\
Quantum Conv. layer2             &             &         &        &      \\ 
Quantum Pooling layer2 & & & & (1, 1, 2) \\ \hline
Partial input2                   & (1, 1, 16)  &         &        &               \\ \hline
Quantum Conv. layer1             & (1, 1, 16)  &         &        &               \\
Quantum Pooling layer1 & & & & \\
Quantum Conv. layer2             &             &         &        &      \\ 
Quantum Pooling layer2 & & & & (1, 1, 2) \\ \hline
Partial input3                   & (1, 1, 16)  &         &        &               \\ \hline
Quantum Conv. layer1             & (1, 1, 16)  &         &        &               \\
Quantum Pooling layer1 & & & & \\
Quantum Conv. layer2             &             &         &        &      \\ 
Quantum Pooling layer2 & & & & (1, 1, 2) \\ \hline
Partial input4                   & (1, 1, 16)  &         &        &               \\ \hline
Quantum Conv. layer1             & (1, 1, 16)  &         &        &               \\
Quantum Pooling layer1 & & & & \\
Quantum Conv. layer2             &             &         &        &   \\ 
Quantum Pooling layer2 & & & & (1 ,1 ,2) \\ \hline
Partial input5                   & (1, 1, 56)  &         &        &               \\ \hline
Classical 1D Conv. layer1        & (1, 1, 56)  & 1 / 8   & 30 / 2 & (1, 8, 24)    \\
Classical 1D Conv. layer2        & (1, 8, 24)  & 8 / 16  & 20 / 2 & (1, 16, 3)    \\ \hline
\begin{tabular}[c]{@{}l@{}}Concatenate outputs\\ from quantum and classical Conv layers\end{tabular} &
  (1, 1, 56) &
   &
   &
   \\ \hline
Classical fully connected layer1 & (1, 1, 56)  &         &        & (1, 1, 18)    \\
Classical fully connected layer2 & (1, 1, 18)  &         &        & (1, 1, 9)     \\
Classical fully connected layer3 & (1, 1, 9)   &         &        & (1, 1, 2)     \\ \hline
Output                           &             &         &        & (1, 2)        \\ \hline
\bottomrule[1.0pt]
\end{tabular}
\end{table}

\clearpage
\section{Statistical test results for classification performances between hybrid quantum-classical and baseline models.}
\setcounter{table}{0}
\renewcommand{\thetable}{C\arabic{table}}

\begin{table}[ht!]
\caption{t-test results for differences between the hybrid and the baseline models.}
\centering
\renewcommand{\arraystretch}{1.3}
\begin{tabular}{lcc}
\hline
 & Baseline model & \begin{tabular}[c]{@{}c@{}}Hybrid model 1\\ (with single quantum part)\end{tabular} \\ \hline
Mean                         & 0.5228                     & 0.562  \\
Variance                     & 0.0002                     & 0.0008 \\
Observations                 & 116                        & 116    \\
Pearson Correlation          & 0.313536158                &        \\
Hypothesized Mean Difference & 0                          &        \\
df                           & 115                        &        \\
t Stat                       & -15.64308714               &        \\
P(T\textless{}=t) one-tail   & \textit{\textbf{1.50E-30}} &        \\
t Critical one-tail          & 1.65821177                 &        \\
P(T\textless{}=t) two-tail   & \textit{\textbf{2.99E-30}} &        \\
t Critical two-tail          & 1.980807476                &        \\ \hline
 & Baseline model & \begin{tabular}[c]{@{}c@{}}Hybrid model 2\\ (with two quantum parts)\end{tabular}   \\ \hline
Mean                         & 0.5228                     & 0.5745 \\
Variance                     & 0.0002                     & 0.0006 \\
Observations                 & 116                        & 116    \\
Pearson Correlation          & -0.084718363               &        \\
Hypothesized Mean Difference & 0                          &        \\
df                           & 115                        &        \\
t Stat                       & -18.6588469                &        \\
P(T\textless{}=t) one-tail   & \textit{\textbf{6.94E-37}} &        \\
t Critical one-tail          & 1.65821177                 &        \\
P(T\textless{}=t) two-tail   & \textit{\textbf{1.39E-36}} &        \\
t Critical two-tail          & 1.980807476                &        \\ \hline
 & Baseline model & \begin{tabular}[c]{@{}c@{}}Hybrid model 3\\ (with four quantum parts)\end{tabular}  \\ \hline
Mean                         & 0.5228                     & 0.5806 \\
Variance                     & 0.0002                     & 0.0007 \\
Observations                 & 116                        & 116    \\
Pearson Correlation          & 0.029713522                &        \\
Hypothesized Mean Difference & 0                          &        \\
df                           & 115                        &        \\
t Stat                       & -21.21479802               &        \\
P(T\textless{}=t) one-tail   & \textit{\textbf{7.29E-42}} &        \\
t Critical one-tail          & 1.65821177                 &        \\
P(T\textless{}=t) two-tail   & \textit{\textbf{1.46E-41}} &        \\
t Critical two-tail          & 1.980807476                &        \\ \hline
\end{tabular}
\end{table}

\clearpage
\begin{table}[ht]
\centering
\renewcommand{\arraystretch}{1.3}
\caption{t-test results for differences between three model conditions in the hybrid model.}
\begin{tabular}{lcc}
\hline
 &
  \begin{tabular}[c]{@{}c@{}}Hybrid model 1\\ (with single quantum part)\end{tabular} &
  \begin{tabular}[c]{@{}c@{}}Hybrid model 2\\ (with two quantum parts)\end{tabular} \\ \hline
Mean                         & 0.562        & 0.5745 \\
Variance                     & 0.0008       & 0.0006 \\
Observations                 & 116          & 116    \\
Pearson Correlation          & 0.163658677  &        \\
Hypothesized Mean Difference & 0            &        \\
df                           & 115          &        \\
t Stat                       & -3.962273896 &        \\
P(T\textless{}=t) one-tail   & \textit{\textbf{6.46E-05}}     &        \\
t Critical one-tail          & 1.65821177   &        \\
P(T\textless{}=t) two-tail   & \textit{\textbf{1.29E-04}}     &        \\
t Critical two-tail          & 1.980807476  &        \\ \hline
 &
  \begin{tabular}[c]{@{}c@{}}Hybrid model 1\\ (with single quantum part)\end{tabular} &
  \begin{tabular}[c]{@{}c@{}}Hybrid model 3\\ (with four quantum parts)\end{tabular} \\ \hline
Mean                         & 0.562        & 0.5806 \\
Variance                     & 0.0008       & 0.0007 \\
Observations                 & 116          & 116    \\
Pearson Correlation          & 0.302224092  &        \\
Hypothesized Mean Difference & 0            &        \\
df                           & 115          &        \\
t Stat                       & -6.341725888 &        \\
P(T\textless{}=t) one-tail   & \textit{\textbf{2.31E-09}}     &        \\
t Critical one-tail          & 1.65821177   &        \\
P(T\textless{}=t) two-tail   & \textit{\textbf{4.62E-09}}     &        \\
t Critical two-tail          & 1.980807476  &        \\ \hline
 &
  \begin{tabular}[c]{@{}c@{}}Hybrid model 2\\ (with two quantum parts)\end{tabular} &
  \begin{tabular}[c]{@{}c@{}}Hybrid model 3\\ (with four quantum parts)\end{tabular} \\ \hline
Mean                         & 0.5745       & 0.5806 \\
Variance                     & 0.0006       & 0.0007 \\
Observations                 & 116          & 116    \\
Pearson Correlation          & 0.638690571  &        \\
Hypothesized Mean Difference & 0            &        \\
df                           & 115          &        \\
t Stat                       & -3.085865669 &        \\
P(T\textless{}=t) one-tail   & \textit{\textbf{0.001}}        &        \\
t Critical one-tail          & 1.65821177   &        \\
P(T\textless{}=t) two-tail   & \textit{\textbf{0.003}}        &        \\
t Critical two-tail          & 1.980807476  &        \\ \hline
\end{tabular}
\end{table}

\clearpage
\section{The classification performance of the baseline model with the Long shot-term memory (LSTM) structure}
\setcounter{table}{0}
\renewcommand{\thetable}{D\arabic{table}}

{To investigate the classification performance of the baseline model including the recurrent neural network (RNN) structure, we conducted additional experiments on the new baseline model with the single LSTM layer. The model architecture of the new baseline model is depicted in Figure D.1. The new baseline model has 13,002 trainable parameters, which is similar to the number of trainable parameters in the original baseline model (11,605 parameters) used in this study (Table D1). All experimental conditions remain the same (0.0001 learning rate, cross-entropy loss function with class weight, single batch, and 100 epochs). An independent two-sample t-test was performed to compare the classification performance between the new and the original baseline models. The results showed similar averaged balanced-accuracy values (Table D2).}

\begin{figure}[ht!]
\caption{The baseline model architecture with the single LSTM layer.}
\centering
\tmpframe{\includegraphics[width=0.99\textwidth]{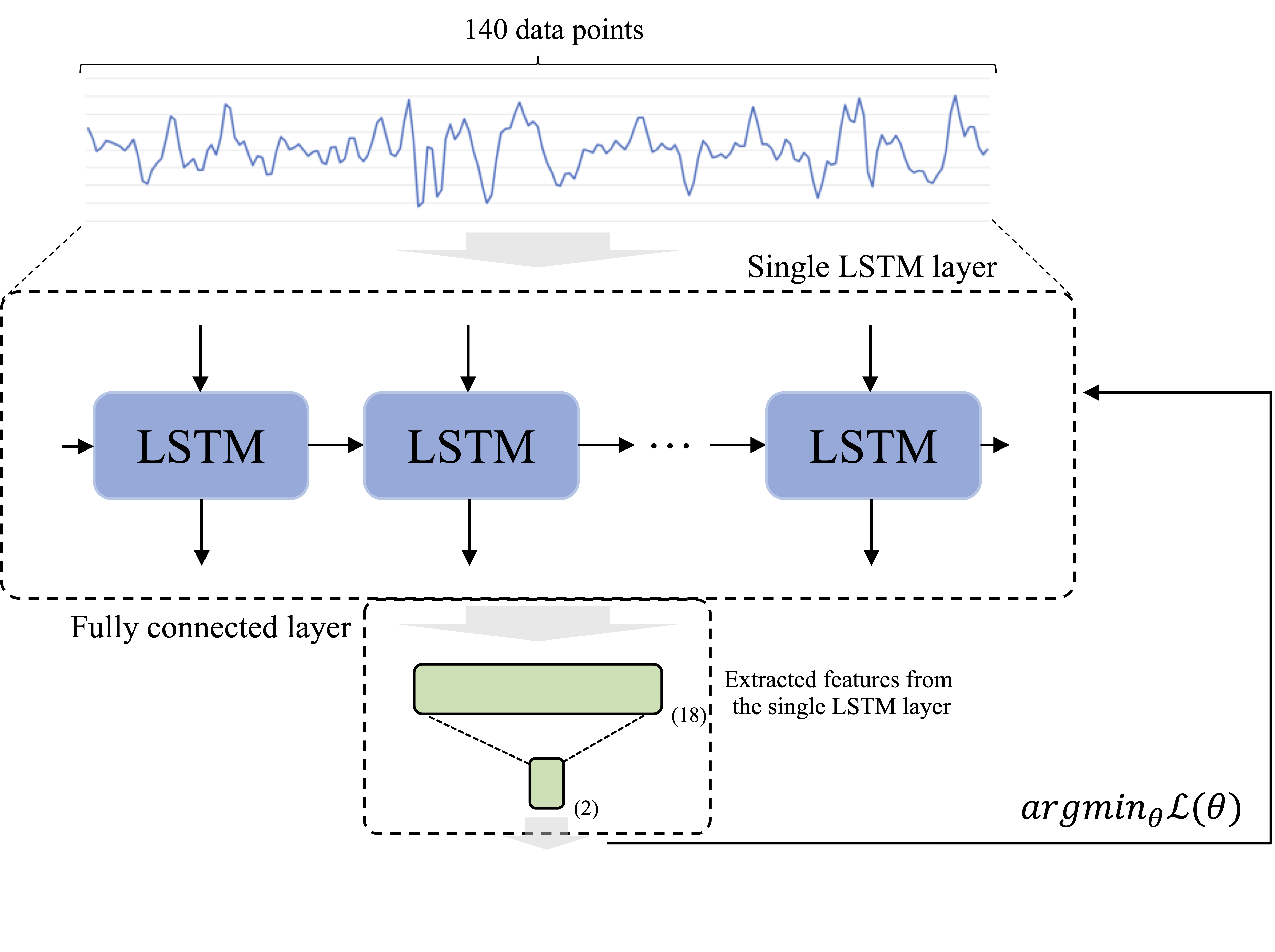}}
\label{figd1}
\end{figure}

\pagebreak
\begin{table}[ht]
\caption{The number of trainable parameters in the baseline model with the LSTM layer and baseline model that was used in this study.}
\centering
\renewcommand{\arraystretch}{1.3}
\begin{tabular}{ccc}
\toprule[1.0pt]
\hline
No. & Model conditions & \# of parameters \\ \hline
1   & Baseline model   & 11,605           \\ \hline
2 & \begin{tabular}[c]{@{}c@{}}Baseline model \\ with the single LSTM layer\end{tabular} & 13,002 \\ \hline
\bottomrule[1.0pt]
\end{tabular}
\label{tab:table}
\end{table}

\begin{table}[ht]
\caption{t-test results of the classification performance between the baseline model with the single LSTM layer and the baseline model that was used in this study.}
\centering
\renewcommand{\arraystretch}{1.3}
\begin{tabular}{ccc}
\toprule[1.0pt]
\hline
 & \begin{tabular}[c]{@{}c@{}}Baseline model\\ (with the single LSTM layer)\end{tabular} & \begin{tabular}[c]{@{}c@{}}Baseline model\\ (original)\end{tabular} \\ \hline
Mean                       & 0.5184                   & 0.5227 \\
Variance                   & 0.0003                   & 0.0002 \\
Observations               & 116                      & 116    \\
t Stat                     & -2.116                   &        \\
P(T\textless{}=t) one-tail & \textit{\textbf{0.0183}} &        \\
t Critical one-tail        & 1.658                    &        \\
P(T\textless{}=t) two-tail & \textit{\textbf{0.0365}} &        \\
t Critical two-tail        & 1.981                    &        \\ \hline
\bottomrule[1.0pt]
\end{tabular}
\label{tab:table}
\end{table}

\end{appendices}

\end{document}